\documentclass[lettersize,journal]{IEEEtran}
\usepackage{amsmath,amsfonts}
\usepackage{algorithmic}
\usepackage{algorithm}
\usepackage{array}
\usepackage[caption=false,font=normalsize,labelfont=sf,textfont=sf]{subfig}
\usepackage{textcomp}
\usepackage{stfloats}
\usepackage{url}
\usepackage{verbatim}
\usepackage{graphicx}
\usepackage{cite}
\usepackage{times}
\usepackage{paralist}
\usepackage{epstopdf}
\usepackage{epsfig}
\usepackage{amssymb}
\usepackage{xcolor}
\usepackage[square,numbers]{natbib}
\usepackage{multirow}
\usepackage{rotating}
\usepackage{stfloats}
\usepackage{ifthen}
\usepackage{times}
\usepackage{dsfont}
\usepackage{booktabs}
\usepackage{soul}
\usepackage{wrapfig}
\usepackage{hyperref}

\usepackage[font=small]{caption}

\newcommand{\eg}{\textit{e.g.}}
\newcommand{\relu}{\mathrm{ReLU}}
\newcommand{\calI}{\mathcal{I}}

\newcommand{\CUT}[1]{}

\newcommand{\calT}{\mathcal{T}}
\newcommand{\calL}{\mathcal{LP}}
\newcommand{\calA}{\mathcal{A}}

\newcommand{\real}{\mathbb{R}}

\newcommand{\red}[1]{{#1}}
\newcommand{\rred}[1]{{#1}}
\newcommand{\bblue}[1]{\textcolor{blue}{#1}}

\newcommand{\zcy}[1]{{#1}}
\newcommand{\zcyNOTE}[1]{\textcolor{cyan}{[ZCY:#1]}}

\newcommand{\abc}[1]{{#1}}
\newcommand{\abcn}[1]{{#1}}

\usepackage{colortbl}
\definecolor{mygray}{gray}{.9}

\usepackage{silence}
\begin{document}

\title{Grad-ECLIP: Gradient-based Visual \abc{and Textual} Explanations for CLIP}

\author{Chenyang~Zhao, Kun~Wang, Janet~H.~Hsiao and Antoni~B.~Chan,
\thanks{Chenyang Zhao is with the Department of Computer Science, City University of Hong Kong, and the Division of Social Science, Hong Kong University of Science \& Technology. 
Antoni~B.~Chan (corresponding author)  is with the Department of Computer Science, City University of Hong Kong. Janet~H.~Hsiao is with the Division of Social Science and Department of Computer Science \& Engineering, Hong Kong University of Science \& Technology, and Kun~Wang is with the SenseTime Group Ltd.
E-mail: zhaocy2333@gmail.com, abchan@cityu.edu.hk.}
\thanks{
}}

\markboth{Journal of \LaTeX\ Class Files,~Vol.~X, No.~X, XX~XXXX}%
{Shell \MakeLowercase{\textit{et al.}}: A Sample Article Using IEEEtran.cls for IEEE Journals}


\maketitle

\begin{abstract}
	Significant progress has been made in the improvement and downstream applications of the Contrastive Language-Image Pre-training (CLIP) vision-language model, while less attention has been paid to the interpretation of CLIP. We propose a Gradient-based visual \abc{and textual} Explanation method for CLIP (Grad-ECLIP), which interprets the matching result of CLIP for a specific input image-text pair. By decomposing the encoder's architecture and identifying the relationship between matching similarity and intermediate spatial features, Grad-ECLIP generates effective heat maps that reveal the impact of image regions or words on the CLIP results. Unlike previous Transformer interpretation methods that focus on utilizing self-attention maps, which are typically extremely sparse in CLIP, we produce high-quality visual explanations by applying channel and spatial weights to token features. 
	Qualitative and quantitative evaluations verify the effectiveness and superiority of Grad-ECLIP compared with the state-of-the-art methods. Finally, a series of analyses are conducted based on our visual and textual explanation results, from which we explore the working mechanism of image-text matching,  the strengths and limitations in attribution identification of CLIP, \abc{and the relationship between the concreteness/abstractness of a word and its usage in CLIP.}
	The code of Grad-ECLIP is available here: https://github.com/Cyang-Zhao/Grad-Eclip.
\end{abstract}


\vspace{-0.1cm}
\section{Introduction}
\label{sec:intro}
\vspace{-0.1cm}

\CUT{
Recently, by learning the representations for matching caption text and its corresponding image, the Contrastive Language-Image Pre-training (CLIP)  model \cite{radford2021clip} has introduced a simple and effective dual-encoder 
pre-training paradigm for the interaction between natural language processing and computer vision. 
CLIP significantly improves the performance on various downstream tasks, such as classification \cite{changpinyo2021conceptual,cha2022domain}, retrieval \cite{luo2022clip4clip}, and segmentation \cite{wang2022cris,xu2022groupvit}, with zero-shot and fine-tuning methodologies. Inspired by CLIP, multi-modal pre-training has been further developed by exploring different perspectives, including unifying vision-language understanding and generation \cite{yu2022coca,li2022blip}, prompt design \cite{zhou2022learning,chen2022prompt}, and region-aware enhancement \cite{li2020oscar,wang2023position,zhong2022regionclip}. 
Although researchers devote many efforts to improving multi-modal pre-training or exploring the usages in downstream tasks, less attention has been focused on the interpretation or explanation of CLIP. 
}

\red{Recently,  Vision-Language Pre-training (VLP) models have bridged the gap between the low-level visual information in pixels and high-level conceptual knowledge embedded in language. By leveraging self-supervised objectives on web-scale data, VLP models learn an aligned understanding of both vision and language, and can flexibly perform applications on various tasks, such as classification, image-text retrieval, visual questioning answering (VQA), image captioning, and serve as foundational technology for the visual inputs to generative AI systems. The evolution of VLP has been marked by several influential architectures. The Contrastive Language-Image Pre-training (CLIP) \cite{radford2021clip} model, which employs a dual-encoder architecture trained with a contrastive objective to facilitate image-text matching, is a notable milestone. CLIP significantly improves performance on various downstream tasks \cite{changpinyo2021conceptual,cha2022domain,luo2022clip4clip,wang2022cris,xu2022groupvit}, utilizing both zero-shot and fine-tuning methodologies.}  Inspired by CLIP, VLP has been further developed by exploring different perspectives, including unifying vision-language understanding and generation \cite{yu2022coca,li2022blip}, prompt design \cite{zhou2022learning,chen2022prompt}, and region-aware enhancement \cite{li2020oscar,wang2023position,zhong2022regionclip}. 
\red{Although researchers devote many efforts into improving multi-modal pre-training or exploring the usages in downstream tasks, less attention has been focused on the interpretation or explanation of the learned vision and language matching process.}

\begin{figure}
	\begin{center}
		\includegraphics[width=\linewidth]{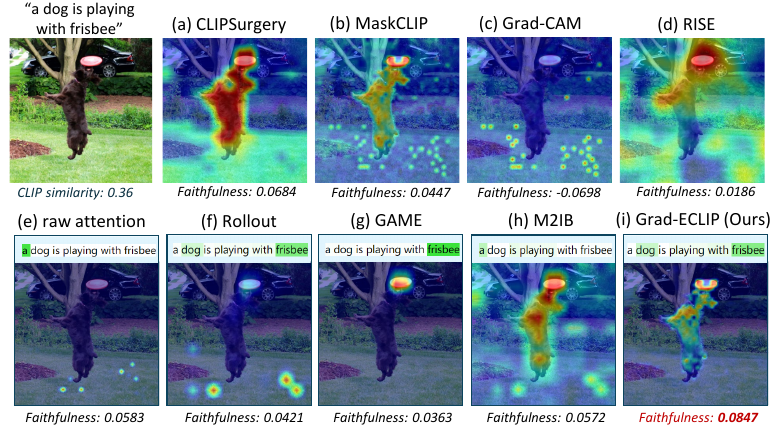}
	\end{center}
	\vspace{-0.3cm}
	\caption{Visual \abc{and textual} explanations of CLIP for the image with the text ``A dog is playing with frisbee'' using (a) CLIPSurgery \cite{li2023clipsurgery}; (b) MaskCLIP \cite{zhou2022extract}; (c) Grad-CAM \cite{selvaraju2017grad}; (d) RISE \cite{petsiuk2018rise};  (e) raw attention in the last layer; (f) Rollout \cite{abnar2020quantifying}; (g) GAME \cite{chefer2021generic}; (h) M2IB \cite{wang2024visual}; and (i) Our Grad-ECLIP. \zcy{For (e) to (i), textual explanations on the sentence are shown, where the degree of green color represents the word importance. Other methods (a-d) are not applicable to text.  \red{The faithfulness (IMD) of each method on the image is shown under the heat map of each method (higher is better).}}
	}
	\vspace{-0.5cm}
	\label{fig:image1}
\end{figure}

\red{Previous visual explanation works have considered interpreting the transformer architecture used by VLP models.}
Attention Rollout \cite{abnar2020quantifying} generates explanations by aggregating attention maps computed along the forward pass of the model. 
Relevance-based methods \cite{chefer2021transformer,chefer2021generic} apply Layer-wise Relevance Propagation (LRP) \cite{bach2015pixel} and also rely on the attention mechanism in the model architecture. 
Since Rollout and many LRP variants are class-agnostic, Transformer interpretability \cite{chefer2021transformer} and Generic Attention-Model Explainability (GAME) \cite{chefer2021generic} build class-specific relevance-based explanations using the self-attention or co-attention. 
However, \red{just treating VLP models as a vision transformer} and generating explanations based on self-attention sometimes leads to \red{low-faithfulness and confusing visualization results} because of sparse attention maps (see Fig.~\ref{fig:image1}e-g). \red{Figure~\ref{fig:image1} shows the visual explanation heat maps on CLIP with the evaluated IMD faithfulness, where the value represents an average influence on the image-text similarity score within 100 steps when deleting and inserting image pixels based on the heat maps (see \S\ref{sec:del_ins} for more details about the IMD metric). Higher IMD values are better, indicating better fidelity to the actual important regions.}

\red{The vision-language matching interpretation is explored by ECLIP \cite{li2022exploring} and CLIPSurgery \cite{li2023clipsurgery} with CLIP by computing an image-text similarity map. They attempt to}
solve the counter-intuitive problem that background patch features get higher similarity with the text feature than the foreground. However, to obtain reasonable similarity maps, new additional projection layers or changing the structure of the original CLIP are required. Although the parameters of CLIP encoders are frozen, learning more black-box parameters with extra data or modifying the original model architecture makes the explanation less interpretable.
MaskCLIP \cite{zhou2022extract} also provides a technique to calculate class-specific image-text similarity map \red{by passing the value features of the last attention layer through later linear layers as image patch features. Although the similarity maps from CLIPSurgery and MaskCLIP are able to localize the concept in the text (see Fig.~\ref{fig:image1}a-b), lower faithfulness results come from a noisy background and confusingly highlight points on the location unrelated to the explained target matching text.} The disadvantage of these similarity-map methods is that they are only forward processing, and the attended features are not necessarily used in the final prediction.

To better focus on the discriminative features used in the prediction, gradient-based methods with class-activation maps (CAM) \cite{zeiler2014visualizing}, such as Grad-CAM \cite{selvaraju2017grad}, Layer-CAM \cite{jiang2021layercam}, and FullGrad \cite{NEURIPS2019_80537a94}, consider the gradient of the prediction with respect to features from a CNN layer as weights, and 
locates the class-specific discriminative regions by weighted aggregation of the feature maps. \red{Fig.\ref{fig:image1}c shows the faithfulness and visualization} when adapting Grad-CAM on CLIP,
where the cosine similarity of the image-text pair is adopted as the prediction and the gradients are calculated w.r.t.~the patch tokens from the ViT layers. 
Since there are no gradients w.r.t.~the patch tokens in the final layer because they are not involved in the calculation of the matching score, feature outputs from the penultimate layer of ViT are adopted.
The results of Grad-CAM do not well explain CLIP, and suffer from the same problem of highlighting unrelated points as MaskCLIP, suggesting that the layer features of ViT are unsuitable for CAM methods.  

\red{In this paper, we explore a more effective way to interpret the image-text matching in VLP models. Using the representative CLIP dual-encoders as the paradigm architecture, we analyze how CLIP obtains the final feature embedding and derive the relationship between the image and text embedding similarity score and intermediate features via a series of approximations.}
\red{We first make the derivation based on the more widely used ViT-based CLIP, which is suitable for both the transformer-based image and text encoders. Then, we give the adaptation of the CNN-based architecture.}
Based on the CAM principle, we then propose a novel gradient-based visual explanation method for CLIP (Grad-ECLIP), which generates the importance heat map by aggregating the intermediate \abc{image} features with result-related \emph{channel} and \emph{spatial} importance weights.
Our proposed method  
uses the gradients of the image-text matching score 
w.r.t.~the
attention layer as the importance for feature channels. For the spatial importance, because the softmax attention typically yields sparse attention maps, we propose a loosened attention map for computing the spatial importance, which can better reflect the importance of more regions, as compared to directly using the strict softmax attention.
Then our Grad-ECLIP explanation map is calculated with the \textit{values} in the attention layer as the feature map, weighted by the channel and spatial importances. 
\abc{The same method used to generate the explanation of the image encoder can also be applied to the text encoder to obtain a textual explanation for CLIP.}
Note that Grad-ECLIP is result-specific and is suitable for both the image and text encoders, i.e., the visual explanation on image is text-specific and the \abc{textual explanation of a sentence}  
is image-specific. 
\red{As the example shows in Fig.~\ref{fig:image1}i, Grad-ECLIP produces the best faithfulness result on this example image-text pair. The} heat map on the image shows the important region when matching the image with the specific text ``a dog is playing with frisbee'',
while the degree of green color on the sentence represents the \abc{important words}, 
where the most important words ``dog'' and ``frisbee'' correspond to the highlighted regions in the image heat map.

In the experiments, we conduct both qualitative evaluation by visualization of explanation maps and quantitative evaluation compared with other 
types of explanation methods, and show the superiority of our proposed Grad-ECLIP. \zcy{Moreover, in the qualitative evaluation, we demonstrate the generalizability of Grad-ECLIP by applying our method on diverse datasets of different domains and showing it is applicable to both ViT and CNN-based CLIP, as well as the ViT classifier and other transformer-based vision language models like BLIP \cite{li2022blip}.}
Then, using Grad-ECLIP, we further conduct a visualization-based analysis on CLIP, and reveal working mechanisms and advantages/limitations of the CLIP model, \abc{including the type of attributes and the concreteness/abstractness of words used by CLIP}. We hope our proposed method can be helpful for researchers to explore more properties of vision-language models like CLIP, \abc{as well as understand their current limitations and how this may affect downstream tasks}.

\zcy{Finally, we also present an application of Grad-ECLIP to fine-tuning the CLIP model to boost the fine-grained understanding. Since the ViT-based CLIP model has been shown to have limitations in producing dense representations, due to the pretraining focusing on the whole image-text matching \cite{zhong2022regionclip,wang2023position, kim2023region,wu2023clipself},
	we propose to generate detailed region-phrase corresponding pairs via Grad-ECLIP so as to enhance the fine-grained understanding of the CLIP model during fine-tuning. Experiments with zero-shot region classification and downstream open-vocabulary detection application show that the Grad-ECLIP-enabled fine-tuning is effective.}


In summary, the contributions of this paper are fivefold: 
\begin{compactenum}
	 \item \red{ We explore the explanation of image-text matching in VLP models by investigating Grad-ECLIP, a gradient-based visual and textual explanation approach for CLIP dual-encoders to produce high-quality result-specific heat maps for explaining the matching of image-text pairs.}
	
	\item We demonstrate the superiority of the proposed Grad-ECLIP with comprehensive evaluations 
	comparing with the state-of-the-art explanation methods for Transformers and CLIP.
	
	\item \red{We show the generalizability of Grad-ECLIP by verifying the applicability on both ViT and CNN-based CLIP, other transformer-based VLP models like BLIP, as well as ViT classifier, and presenting explanations on datasets of different domains.}
	
	\item By using Grad-ECLIP, we explore the properties of CLIP, and reveal the model's ability of concept decomposition and addability, strengths and weaknesses in attribution identification\zcy{, as well as the relationship between word \abc{usage} and concreteness in the image and text matching}. 
	
\end{compactenum}

\zcy{A preliminary version of our work appears in \cite{zhao2024gradient}. The extensions over the conference version are as follows. First, we provide a more detailed derivation of Grad-ECLIP, 
	based on the transformer model. Second, we enrich the qualitative evaluation and verify the generalizability of Grad-ECLIP by adding: (1) the visualization on diverse datasets of different domains; (2) application on ViT-based classifier and other vision language models; (3) adaptation with CNN-based CLIP. Third, we include new ablation studies for the Grad-ECLIP design, including: (1) the effect of the proposed loosen spatial weight; (2) the influence of the number of layers involved in the calculation; (3) the influence of multi-attention heads on the visual explanation results. Fourth, we add \red{quantitative experiments for the analysis of CLIP} and a new exploration about the relationship between word concreteness and word usage in image-text matching with CLIP. \red{Moreover, we conduct a user study to evaluate the trustability of the visual explanations.}
	}

\zcy{The remainder of this paper is organized as follows. The related works about CLIP and explainability in computer vision are briefly reviewed in \S\ref{sec:related}. Grad-ECLIP is introduced in \S\ref{sec:method}, and the experiment results are presented in \S\ref{sec:exp}, including qualitative evaluations, quantitative evaluations, and ablation studies. Finally, we perform analysis of CLIP based on the proposed Grad-ECLIP in \S\ref{sec:analysis_clip}. 
	}

\vspace{-0.2cm}
\section{Related Works}
\label{sec:related}
\vspace{-0.1cm}

\zcy{We first briefly review the 
	CLIP
	model, and then discus different types of visual explanation methods in computer vision. 
	}

\vspace{-0.1cm}
\subsection{Contrastive language-image pre-training}

Many multi-modal works have been developed and focus on the interaction of computer vision and natural language processing, such as text-image retrieval \cite{wang2019camp}, image captioning \cite{xu2015show}, visual question answering \cite{antol2015vqa}, and visual grounding \cite{plummer2015flickr30k}.
Contrastive language-image pre-training (CLIP) performs contrastive learning on very large-scale web-curated image-text pairs. It shows promising pre-trained representations with superior zero-shot transfer ability on diverse datasets and impressive fine-tuning performance on various downstream tasks. 
Subsequent works extend and improve CLIP from different aspects: \citep{zhou2022learning,chen2022prompt} improve the aspects of prompt design and optimization; \citep{yu2022coca,li2022blip} unify the vision-language understanding and generation by adding text decoders with image-text cross-attention during pre-training; \citep{li2020oscar,wang2023position,zhong2022regionclip,wu2023clipself} build an alignment between region feature or position information with fine-grained object descriptions. 
Although significant results have been achieved with CLIP and its development, less effort and exploration are focused on its interpretability through visual explanations. In this paper, we propose a novel visual explanation method, which generates high-quality heat maps that reveal the important regions or words used for CLIP's scoring of an image-text pair. 

\vspace{-0.1cm}
\subsection{Explainability in computer vision}

Since visualizing the importance of input features is a straightforward approach to interpret a model, many works visualize the internal representations of CNNs or Transformers with heat maps. Most explanation methods can be categorized as: CAM methods, perturbation methods, Shapley-value methods, or attribution propagation (relevance-based) methods.

CAM methods, such as CAM \cite{zeiler2014visualizing}, Grad-CAM \cite{selvaraju2017grad}, and Grad-CAM++ \cite{chattopadhay2018grad}, generate the explanation heat map from a selected feature layer by linearly aggregating the activation maps with weights that indicate each feature's importance. Grad-CAM computes the weights with global average pooling on the gradients of the model's prediction w.r.t the feature layer. Gradient-free CAMs \cite{ramaswamy2020ablation, wang2020score, wang2020ss} generate weights from the prediction score changes when perturbing features.

Perturbation-based methods \cite{ribeiro2016should, petsiuk2018rise,fong2017interpretable, lundberg2017unified, wagner2019interpretable, lee2021bbam, petsiuk2021black} perturb the input and observe the changes in output scores to determine the importance of input regions. Such black-box methods are intuitive and highly generalizable \abc{to different architectures and tasks}, but computationally intensive. The quality of these methods is often greatly influenced by the type or resolution of the perturbations used.
While having solid theoretical justification, Shapley-value methods \cite{lundberg2017unified} also suffer from large computational complexity. 

Attribution propagation methods recursively decompose the network output into the contribution of early layers, based on the Deep Taylor Decomposition (DTD) \cite{montavon2017explaining}. LRP \cite{bach2015pixel} and its variants \cite{lundberg2017unified,nam2020relative,shrikumar2017learning} propagate relevance from the prediction to the input image based on  DTD and generate class-agnostic explanations, while Contrastive-LRP \cite{gu2019understanding} and SG-LRP \cite{iwana2019explaining} generate class-specific explanations.

\abc{These previous works are mainly proposed for interpreting CNN-based models.
	Due to the introduction of self-attention mechanisms in Transformers,  recent works \cite{qiang2022attcat,xie2022vit,yu2023x} have also looked at visual explanations for the  Transformer architecture.}
\cite{abnar2020quantifying} proposed an Attention flow and Rollout method, which is based on all attention maps in the forward process of the model. Since Rollout is class-agnostic, Transformer interpretability \cite{chefer2021transformer} and GAME \cite{chefer2021generic} build class-specific relevance-based methods for explaining the transformer with the internal attention mechanism. However, we found that the explanation methods relying on attention maps in Transformer cannot generate satisfactory results with CLIP, possibly because of the sparse attention patterns on the $\mathrm{softmax}$ map.
The recent M2IB \cite{wang2024visual} applies the information bottleneck principle to CLIP, which develops an optimization objective to find the compressed representations for both image features and text features. However, a series of hyperparameters are adopted during the optimization, which limits the generalization in practical applications.

Finally, existing approaches for explaining CLIP \cite{li2022exploring, li2023clipsurgery, zhou2022extract}, 
which use the cosine similarity map between the image local features and the text features as the explanation map, have the disadvantage that they are only based on the \emph{forward (bottom-up) process} and thus the attended features are not necessarily used in the final prediction. In contrast, we propose Grad-ECLIP as an effective approach to interpret CLIP, which highlights features that have the largest influence on the prediction as measured by the gradient, which is a top-down process.

\begin{figure}
\begin{center}
\includegraphics[width=0.5\textwidth]{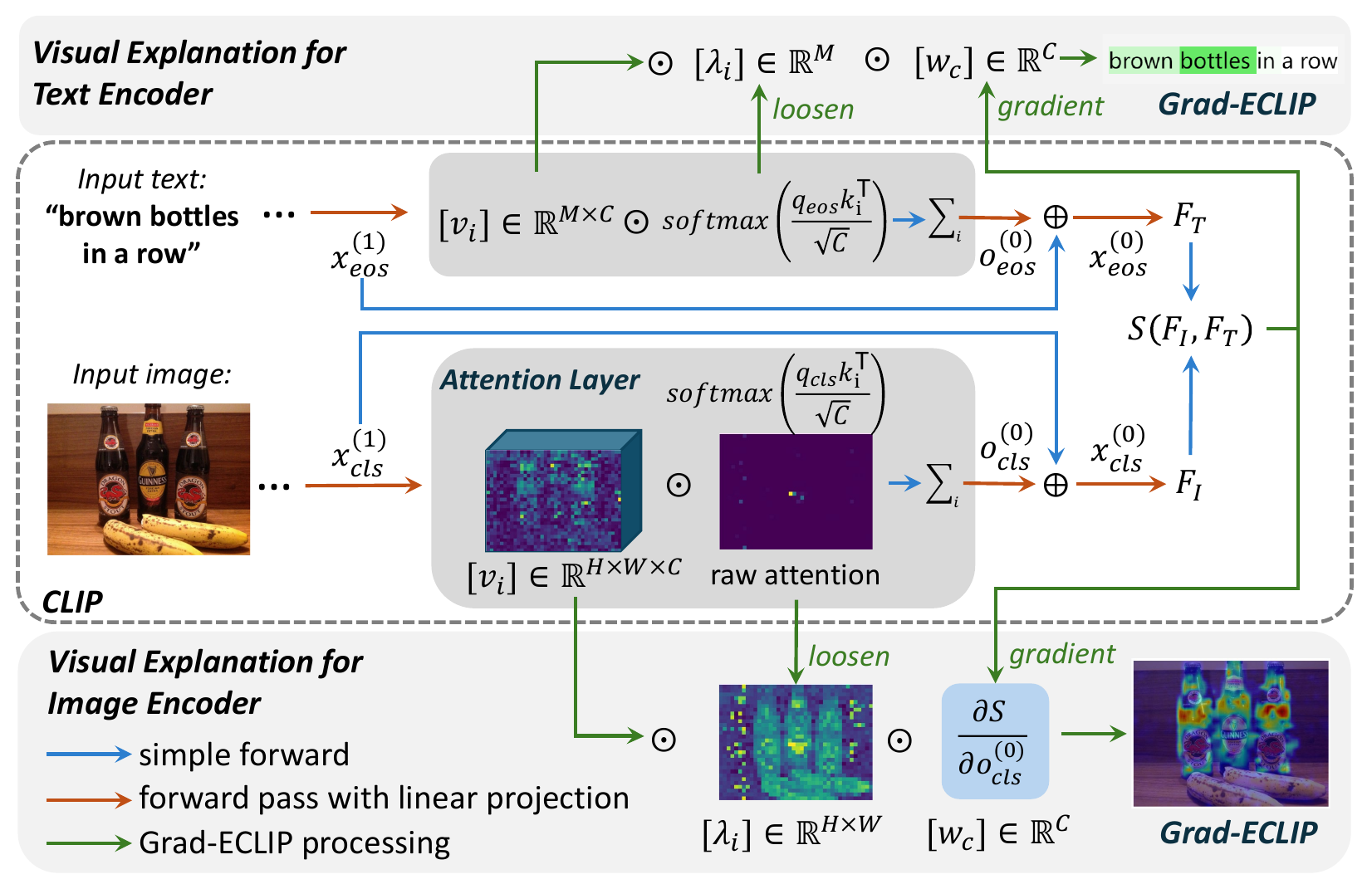}
\end{center}
\vspace{-0.3cm}
\caption{Illustration of Grad-ECLIP. An image-text pair specific visual explanation is generated by weighting and aggregating the \textit{values} as feature map in the attention layer with spatial importance $\lambda_{i}$ and channel importance $w_{c}$. Gradients are propagated to the attention layer output to produce $w_{c}$, and the loosened attention map is applied as $\lambda_{i}$. 
}
\label{fig:framework}
\vspace{-0.5cm}
\end{figure}

\vspace{-0.2cm}
\section{Grad-ECLIP: Gradient-based visual explanation for CLIP}
\label{sec:method}
\vspace{-0.1cm}

Our method serves as a gradient-based visual \abc{and textual} explanation for interpreting the CLIP matching performed on image-text pairs. We start with a brief introduction of preliminaries. Then, by decomposing the layers of the transformer and exploring the relationship between the final output and intermediate features, we derive our gradient-based explanation for CLIP (Grad-ECLIP).

\vspace{-0.2cm}
\subsection{Preliminaries}
\label{sec:pre_clip}
\vspace{-0.1cm}

CLIP learns both visual and language representations from large-scale 
web-curated image-text pairs. It consists of an image encoder $\calI\left(\cdot\right)$ and a text encoder $\calT\left(\cdot\right)$, which are jointly trained to respectively extract image and text feature embedding in a unified representation space. Given image-text pair $(I,T)$, the matching score between their extracted image feature $F_{I}\in \mathbb{R}^{D}$ and text feature $F_{T}\in \mathbb{R}^{D}$ \abc{(both row vectors)} is:
\begin{align}
S(F_{I}, F_{T}) = \cos(F_{I}, F_{T}) = \tfrac{F_{I}F_{T}^\mathsf{T}}{\left \| F_{I} \right \| \left \| F_{T} \right \| } .
\label{eq:cos_similarity}
\end{align}

\abc{The model is trained using}
contrastive learning on the matching scores, regarding the ground-truth image-text pairs as positive samples and other mismatched pairs as negatives. 
\zcy{In practice, both encoders can be implemented as transformers. 
In \S\ref{sec:grad-eclip}, our method is derived based on the transformer architecture, and thus is suitable for interpreting both ViT-based image and transformer-based text encoders. 
Alternatively, the image encoder could be a CNN-based ResNet followed by an attention pooling layer, which 
is basically the same as an attention layer in the Transformer.
Thus, the proposed Grad-ECLIP is also applicable to the CNN-based CLIP, for which we display the visualization results for CLIP with ResNet \cite{he2016deep} backbone in \S\ref{sec:apply_cnn_clip}. }

\red{For a Transformer that consists of $N$ layers, following convention, we denote $x^{(n)}$ as the input of layer $L^{(n)}$ and output of layer $L^{(n+1)}$, where $n\in\left[0...N\right]$ is the layer index. $x^{(N)}$ is the input of the network, $x^{(1)}$ is the input of the last layer and $x^{(0)}=\calI(x^{(N)})$ is the output of the network. 
The image feature is $F_{I}=\calL(x_{cls}^{(0)})$, where $\calL$ denotes linear projections, and $x_{cls}$ is the feature vector from the $[cls]$ token.
Thus, except for the class token, all the final layer features of the other tokens (image patch tokens) are not used during contrastive learning of CLIP. Therefore, to interpret the $S_{T}(F_{I})$ w.r.t. an image feature map, we explore the relationship between the last layer class token feature $x_{cls}^{(0)} \in \real^C$ and the intermediate spatial feature maps. }

\vspace{-0.2cm}
\subsection{\zcy{Methodology of Grad-ECLIP}}
\label{sec:grad-eclip}
\vspace{-0.1cm}

Here we present our derivation of Grad-ECLIP from the image viewpoint, where the visualization is generated on the input image $I$ and shows important regions related to producing the matching score $S_{T}(F_{I}) \triangleq S(F_{I}, F_{T})$, with the given specific text prompt $T$. The application of Grad-ECLIP from the text viewpoint, where the visualization is generated for the text prompt $T$ given the input image $I$, can be obtained analogously by considering the $[eos]$ token (end of sentence token) from the text encoder, which is analogous to the $[cls]$ token in the image encoder.

\CUT{
For a Transformer that consists of $N$ layers, following convention, we denote $x^{(n)}$ as the input of layer $L^{(n)}$ and output of layer $L^{(n+1)}$, where $n\in\left[0...N\right]$ is the layer index. $x^{(N)}$ is the input of the network, $x^{(1)}$ is the input of the last layer and $x^{(0)}=\calI(x^{(N)})$ is the output of the network. 
The image feature is $F_{I}=\calL(x_{cls}^{(0)})$, where $\calL$ denotes linear projections, and $x_{cls}$ is the feature vector from the $[cls]$ token.
Thus, except for the class token, all the final layer features of the other tokens (image patch tokens) are not used during contrastive learning of CLIP. Therefore, to interpret the $S_{T}(F_{I})$ w.r.t. an image feature map, we explore the relationship between the last layer class token feature $x_{cls}^{(0)} \in \real^C$ and the intermediate spatial feature maps. 
}

As shown in the illustration of Fig.~\ref{fig:framework}, looking closely into the last layer of the network, the image embedding from the visual encoder can be formulated as:
\abcn{
\begin{align}
F_{I} &= \calL(x_{cls}^{(0)})  = \calL(o_{cls}^{(0)}+x_{cls}^{(1)}) \\ 
&\approx \calL(o_{cls}^{(0)}) + \calL(x_{cls}^{(1)}), 
\end{align}
where the approximation is based on assuming linearity of $\calL$\rred{\cite{sarlos2006improved, vu2018random}}.
Noting that $ \calL(x_{cls}^{(t)})  = \calL(o_{cls}^{(t)}+x_{cls}^{(t+1)})$, we  substitute recursively to obtain the approximation:
\begin{align}
F_{I} &\approx \calL(o_{cls}^{(0)}) + \cdots +  \calL(o_{cls}^{(N-1)}) + \calL(x_{cls}^{(N)}) \\
&\triangleq \sum_{t=0}^{N} F_I^{(t)}, 
\label{eqn:aggregate}
\end{align}
where we define $F_I^{(t)} = \calL(o_{cls}^{(t)})$ for $t<N$, and $F_I^{(N)} = \calL(x_{cls}^{(N)})$. Thus from (\ref{eqn:aggregate}), the image feature $F_I$ is approximately an aggregation of features from each layer $F_{I}^{(t)}$.}

\abcn{
The feature vector from each layer is computed from a self-attention operation. For example, for the last layer ($t=0$), 
\begin{align}
F_I^{(0)} = \calL(o_{cls}^{(0)}) = \calL\big(\sum_{i}\mathrm{softmax}(\tfrac{q_{cls}k_{i}^\mathsf{T} }{\sqrt{C}})v_{i}\big),
\label{eq:feature}
\end{align}}
where the output of attention layer ($\calA$) on the class token is
\vspace{-0.1cm}
\begin{align}
o_{cls}^{(0)}=\calA(x^{(1)})[cls]=\sum_{i}\mathrm{softmax}(\tfrac{q_{cls}k_{i}^\mathsf{T} }{\sqrt{C}})v_{i},
\label{eq:out_cls}
\end{align}
and $\calA$ represents the attention layer in the Transformer, $q_{cls}$ is the \textit{query} embedding for the class token, while $k_{i}$ and $v_{i}\in\real^C$ represent the \textit{key} and \textit{value} embeddings at spatial location $i$, with $C$ as their channel dimension.\footnote{We skip the superscript $(0)$ on $\{q,k,v\}$ identifying the layer for  brevity.}
The $\mathrm{softmax}$ operation inside the attention layer measures the weight of the value on each location. 
Multi-heads are usually used in the attention layer to group the channels of $\{q, k, v\}$ into several heads, and (\ref{eq:out_cls}) is operated inside each head with the $\mathrm{softmax}$ calculated over subsets of the channels. Then, the final attention layer output is obtained by concatenating the results of each head together. In practice \abcn{for visualization}, we formulate the $o_{cls}^{(0)}$ with one attention head in the forward pass and operate the $\mathrm{softmax}$ over all channels as in (\ref{eq:out_cls}). 
We discuss the influence of multi-heads on visual explanation in the \S\ref{sec:influence_heads}.  

\abcn{
Assuming that the feature vectors $(F_T, F_I)$ are normalized and 
using (\ref{eqn:aggregate}), the matching score can be approximated as an aggregation of partial scores for feature vectors from each layer, 
\begin{align}
S_{T}(F_{I}) &= \sum_{c}F_{T}[c]F_{I}[c] 
\approx \sum_{c}F_{T}[c] \sum_{t} F_{I}^{(t)}[c]  \\
&=\sum_{t} \Big( \sum_{c}F_{T}[c]  F_{I}^{(t)}[c] \Big) 
\triangleq \sum_t S_T(F_I^{(t)}), 
\label{eqn:scoreaggregate}
\end{align}
where $[c]$ selects the $c$-th channel, and $S_T(F_I^{(t)})$ denotes the score from layer $t$. Next, to calculate the heat map for the contribution of $o_{cls}$ on the partial matching score, we write the partial matching score as a function of its $o_{cls}$. Specifically, looking at the last layer ($t=0$) as an example,}
\begin{align}
S_{T}(F_{I}^{(0)}) &= \sum_{c}F_{T}[c]F_{I}^{(0)}[c] 
\\ &= \sum_{c}F_{T}[c]\calL(o_{cls})[c]^{(0)} \triangleq f(o_{cls}), 
\end{align}
\abc{where we have defined the matching score as a function of $o_{cls}$, i.e., $f(o_{cls})$.
We define the approximation of the matching score as a weighted combination of the channel features in $o_{cls}$, we have}
\begin{align}
f(o_{cls}) \approx \tilde{f}(o_{cls}) \triangleq \sum_{c}w_{c}o_{cls}[c] = 
w o_{cls}^{\mathsf{T}}
\label{eq:appr_s}
\end{align}
\abc{where $w_c$ is the weight for the $c$-th channel, and $w=[w_c]_c \in \real^C$ the corresponding weight vector for all channels.}
\abc{To obtain the channel weights $w$, we aim to match the first derivatives (gradients) of the original matching score $f$ and its approximation $\tilde{f}$, leading to the optimization problem:}
\abc{
\begin{align}
w &= \mathop{\mathrm{argmin}}_{w}\big\| f'(o_{cls})-\tilde{f}'(o_{cls})\big\|^{2} \\
&= \mathop{\mathrm{argmin}}  \big\| \tfrac{\partial f}{\partial o_{cls}} - w \big\|^{2}, 
\end{align}
which has the solution
\begin{align}
w &= \tfrac{\partial f}{\partial o_{cls}} = \tfrac{\partial S_T(F_I)}{\partial o_{cls}} .
\label{eq:w_c}
\end{align}
Therefore, substituting (\ref{eq:out_cls}) and (\ref{eq:w_c}) into (\ref{eq:appr_s}), 
\begin{align}
S_{T}(F_{I}) &\approx \sum_{c} w_{c}o_{cls}[c] \\
&= \sum_{c} \tfrac{\partial S_{T}(F_{I})}{\partial o_{cls}[c]} \sum_{i} \mathrm{softmax}(\tfrac{q_{cls}k_{i}^\mathsf{T} }{\sqrt{C}})v_{ic} \\
&= \sum_{i}  \Big[ \sum_{c} \tfrac{\partial S_{T}(F_{I})}{\partial o_{cls}[c]} \mathrm{softmax}(\tfrac{q_{cls}k_{i}^\mathsf{T} }{\sqrt{C}})  v_{ic}\Big],
\end{align}
\abcn{where $v_{ic}$ is the c-th channel of $v_i$.}
Regarding $[v_{ic}]$ as the intermediate feature map, and $\lambda_{i} = \mathrm{softmax}(\tfrac{q_{cls}k_{i}^\mathsf{T} }{\sqrt{C}})$, then the importance of the $i$-th spatial location is defined as:
\begin{align}
H_{i} = \relu\big( \sum_{c} w_{c}\lambda_{i} v_{ic}\big)
\label{eq:hm}
\end{align}
where $\relu$ means that we only focus on positions that have a positive effect on the final score.
}

The weight $w_{c}$ represents the importance of each feature channel, and $\lambda_{i}$ 
represents the importance of the values at each location via the softmax attention. However, from the visualization, we discover that the output of the $\mathrm{softmax}$ self-attention function is extremely sparse. Important information may be encoded in different locations, but the $\mathrm{softmax}$ only selects the largest activation, which is not appropriate as a spatial weight.
Therefore, we replace the $\mathrm{softmax}$ with a ``loosened'' correlation by applying 0-1 normalization on the similarities $[q_{cls} k_i^{\mathsf{T}}]_i$, i.e., $\lambda_{i}\approx\varPhi (q_{cls}k_{i}^\mathsf{T})$, where $\varPhi$ is the 0-1 normalization function applied over the set of similarities. In the experiments and \S\ref{sec:spatial_weight}, we compare using the loosened $\lambda_{i}$ and without $\lambda_{i}$ to show the effect of spatial weights, qualitatively and quantitatively.

Therefore, with \emph{the channel importance $w_{c}$} and \emph{the spatial importance $\lambda_{i}$}, where
\begin{align} 
w_{c}=\tfrac{\partial S_{T}(F_{I})}{\partial o_{cls}^{(0)}[c]}, \hspace{0.1in} \lambda_{i}=\varPhi (q_{cls}k_{i}^\mathsf{T}),
\label{eq:weight}
\end{align}
we  obtain the proposed Grad-ECLIP explanation map $H = [H_{i}]_i$ \abcn{for the last layer}, where $H_i$ is defined in (\ref{eq:hm}) using the last layer values $v$ as the feature map. 

{\bf Visual and textual explanations:}
For the image encoder, the flattened heat map $H_{i}$ is reshaped and interpolated to the original image's height and width, while for the text encoder, the heat (importance) value on the $i_{th}$ tokens is remapped to the original word position in the sentence. 
Finally, based on the approximation in (\ref{eqn:scoreaggregate}), the final explanation can be  \emph{aggregated over all the layers} by recursively processing each layer \abcn{to obtain its heat map from (\ref{eq:hm}).}
In the experiments, we use the last layer to explain the image encoder, and the last eight layers for interpreting the text encoder. 
The ablation study for the influence of different numbers of layers involved in image and text explanation is shown in \S\ref{sec:influnce_layers}. 

{\bf CNN-based CLIP:}
\zcy{
Our proposed Grad-ECLIP can also be applied to CNN-based CLIP. 
The CNN-based CLIP model is composed of a ResNet backbone followed by an attention pooling. Thus, we can use the final attention layer in the pooling to conduct our explanation, which uses the same implementation as for ViT-based CLIP. The visualizations shown in \S\ref{sec:apply_cnn_clip} verify the effectiveness of Grad-ECLIP on CNN-based CLIP model.}

\begin{figure*}
\vspace{-0.2cm}
\begin{center}
\includegraphics[width=0.85\textwidth]{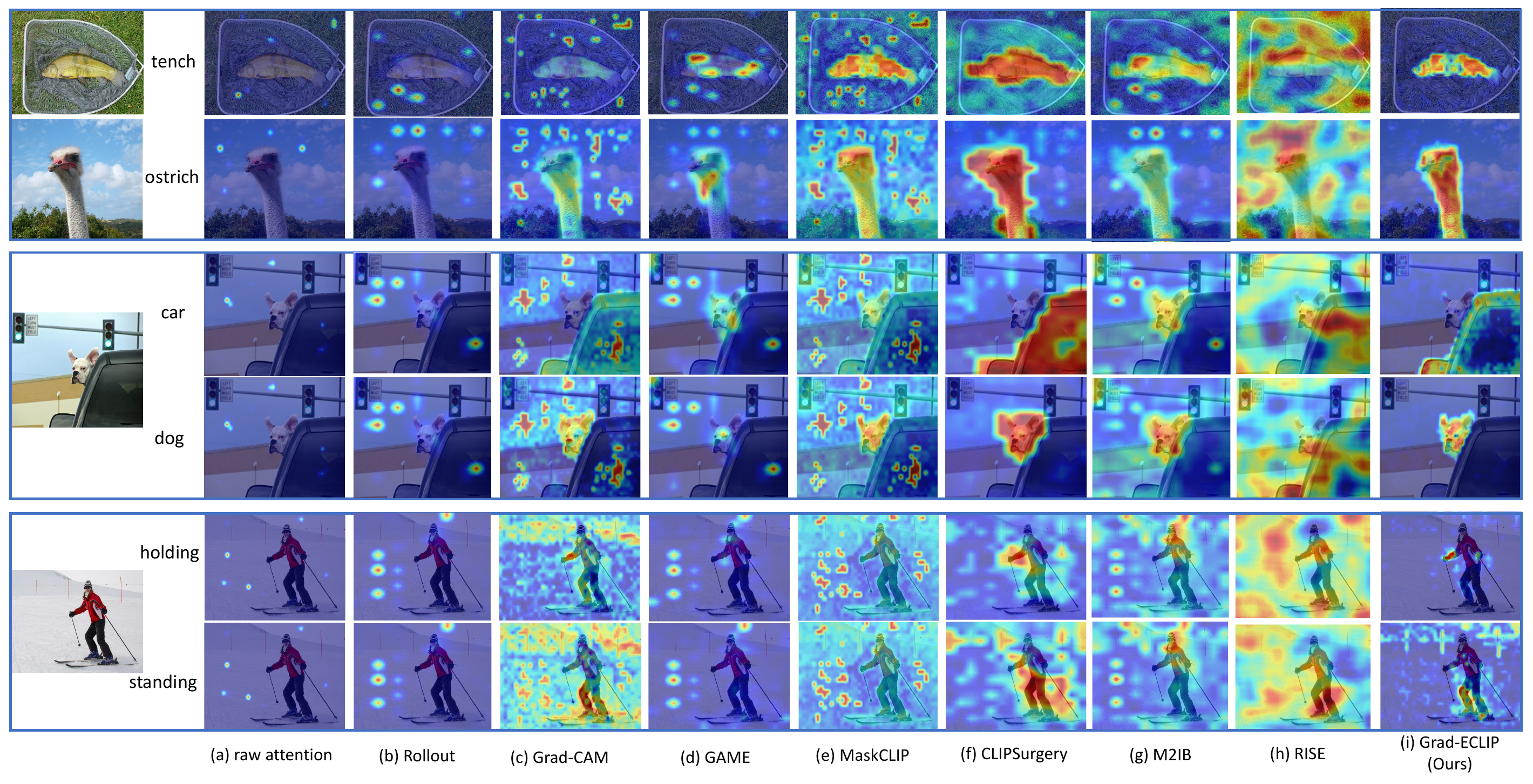}
\end{center}
\vspace{-0.3cm}
\caption{Comparison of heat maps from:
(a) the raw self-attention map in the last ViT \abc{layer}; 
(b) Rollout \cite{abnar2020quantifying}; 
(c) Grad-CAM \cite{selvaraju2017grad}; 
(d) GAME \cite{chefer2021generic}; 
(e) MaskCLIP \cite{zhou2022extract}; 
(f) CLIPSurgery \cite{li2023clipsurgery};
(g) M2IB \cite{wang2024visual};
(h) RISE \cite{petsiuk2018rise}; 
(i) our proposed Grad-ECLIP.	
Visual explanations are provided for the matching score between the image and the specific text prompts, which can be nouns (\eg, car, dog) or verbs (\eg,  holding, standing). From the comparison of visualizations, Grad-ECLIP exhibits superior explanation ability on different types of text prompts. 
}
\label{fig:vis_comp}
\vspace{-0.3cm}
\end{figure*}

\begin{figure*}[!h]
\begin{center}
\includegraphics[width=0.85\textwidth]{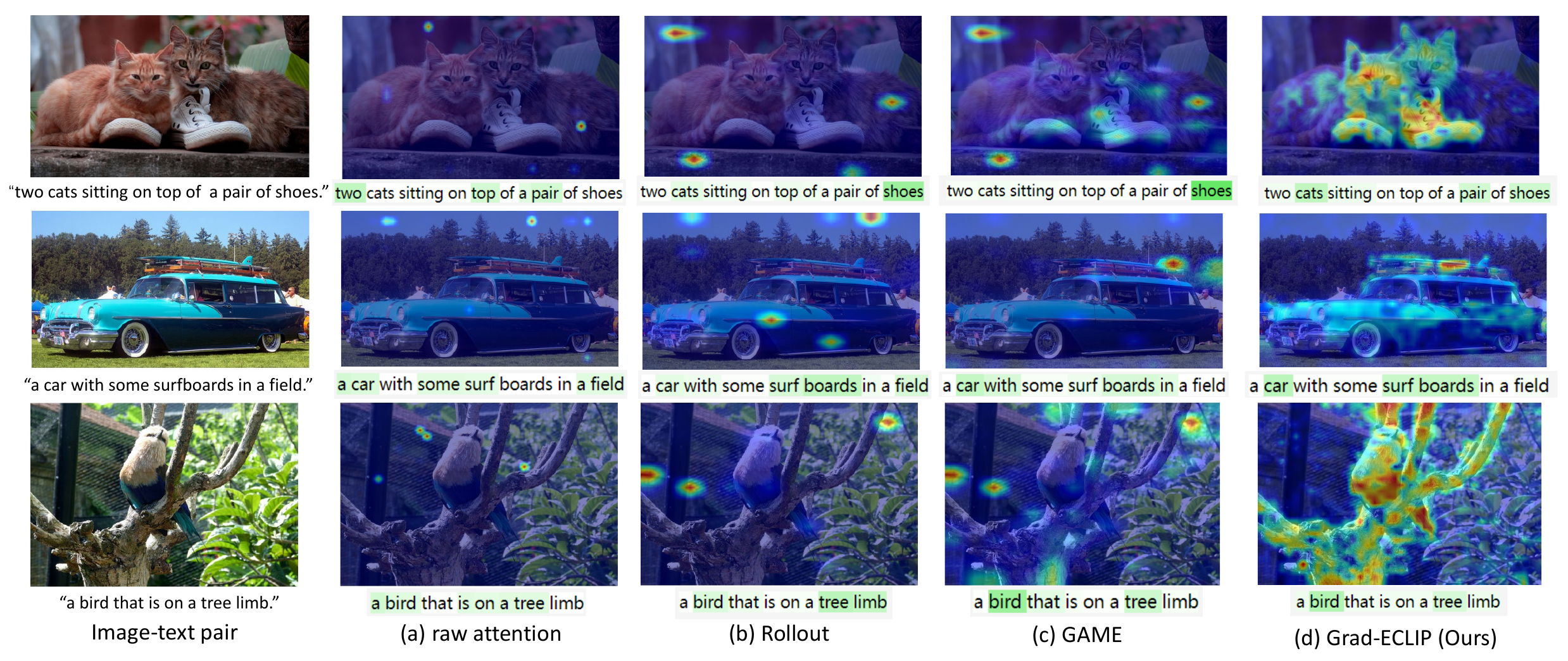}
\end{center}
\vspace{-0.2cm}
\caption{Explanations for image-text pairs from MS COCO using:
by (a) raw self-attention;  Transformer interpretation methods (b) Rollout, (c) GAME; and our method (d) Grad-ECLIP. The importance of words is visualized by the degree of green color.
}
\vspace{-0.2cm}
\label{fig:vis_img_txt}
\end{figure*}

\vspace{-0.2cm}
\section{Experiments with Grad-ECLIP}\label{sec:exp}
\vspace{-0.1cm}

In this section, we conduct experiments on Grad-ECLIP to: 
1) evaluate its visual explanation qualitatively and quantitatively, and compare with the current SOTA methods; 
2) evaluate the processing time;
3) \zcy{conduct ablation studies, including the effect of spatial weight, the involved layers, and attention heads.}

\zcy{Unless otherwise specified,} 
we conducted the experiments with the ViT-B/16 architecture.
%
We compared with representative baseline XAI methods from each category: 
1) attention map-based \textit{Rollout} \cite{abnar2020quantifying}, which takes into account all the attention maps computed along the forward pass, and \textit{raw attention} in the last visual encoder layer, both of which are not result-specific explanations; 
2) classical gradient-based method \textit{Grad-CAM} \cite{selvaraju2017grad}, which takes the image-text similarity as target and calculates the gradients w.r.t. the ViT layer output; 
3) relevance-based \textit{GAME} \cite{chefer2021generic}, which integrates the relevancies and gradients propagated through the network; 
4) cosine-based \textit{MaskCLIP} \cite{zhou2022extract} and \textit{CLIPSurgery} \cite{li2023clipsurgery}, which generate similarity value on each location by the cosine between text feature and processed values as local image features; 
5) M2IB \cite{wang2024visual}, which applies the information bottleneck principle to generate an explanation map for CLIP.
Each baseline is built with different properties and assumptions over the architecture. 
\red{For evaluation of text encoder explanations, we compare Rollout, GAME, and M2IB, and our Grad-ECLIP by generating word importance on the input sentences -- the other methods (Grad-CAM, MaskCLIP, and CLIPSurgery) are excluded since they are only applicable to the image encoder. } 
We also show visualization comparisons with the typical perturbation method RISE \cite{petsiuk2018rise} \red{(and IG \cite{sundararajan2017axiomatic} in the Appendix)}, but did not conduct quantitative comparisons with perturbation and Shapley methods, \red{due to their computational complexity and black-box property. Since the proposed Grad-ECLIP is a white-box method designed for CLIP, we prioritize comparison with other XAI methods designed for CLIP (CLIPSurgery, MaskCLIP, M2IB) or similar transformer architectures (Rollout, GAME), and the classical CAM-based method Grad-CAM, both qualitatively and quantitatively.}

\begin{figure*}[thp]
\centering
\begin{center}
\includegraphics[width=0.85\textwidth]{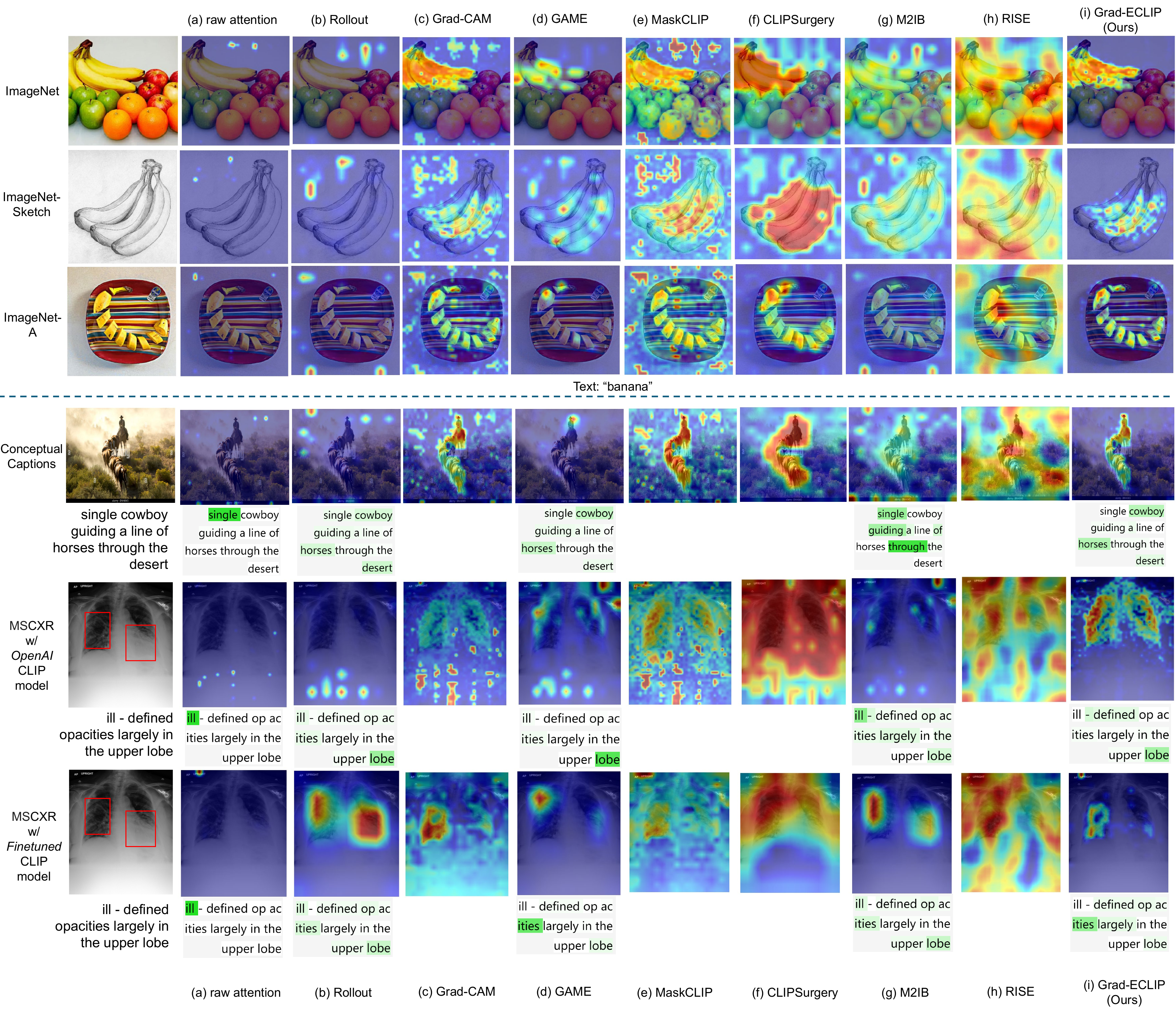}
\end{center}
\vspace{-0.3cm}
\caption{Comparison of visual explanations for different methods on image samples from different image domains: \abc{natural images (ImageNet),  renditions (ImageNet-R), pencil sketch (ImageNet-Sketch), natural adversarial examples (ImageNet-A), web images with captions (Conceptual Captions), and chest X-ray (MSCXR). On MSCXR, explanations are provided for both the OpenAI CLIP model and a fine-tuned version.}}
\vspace{-0.2cm}
\label{fig:vis_domains}
\vspace{-0.2cm}
\end{figure*}

\subsection{Qualitative evaluation}
\label{exp:qual_eval}

\zcy{In this section, we evaluate the proposed Grad-ECLIP qualitatively. 
The comparisons of visual explanations from different types of methods are presented in \S\ref{sec:vis_comp}, and the visual explanations on both image and text encoders with image-sentence pair examples are presented in \S\ref{sec:vis_img_sentence}. Then, we conduct the visualizations of explanations on diverse image domains 
in \S\ref{sec:vis_domain} and adapt the Grad-ECLIP to other Vision Language Models (VLMs) in \S\ref{sec:adapt_vlms}. Finally, we show that the Grad-ECLIP is applicable to CNN-based CLIP with ResNet backbone in \S\ref{sec:apply_cnn_clip}.}

\subsubsection{Comparison of visual explanations}
\label{sec:vis_comp}

We compare the visualizations of raw self-attention, Rollout, Grad-CAM, GAME, MaskCLIP, CLIPSurgery, M2IB, RISE, and our Grad-ECLIP 
in Fig.~\ref{fig:vis_comp} with  images from ImageNet \cite{russakovsky2015imagenet} and MS COCO \cite{lin2014microsoft}. Except for raw attention and Rollout, which are defined to be text-agnostic, the others are all text-specific, so we test the same image with two different text inputs on MS COCO. 
Our Grad-ECLIP  demonstrates a strong ability of generating distinct text-specific heat maps, and gives 
\red{corresponding} explanation of verbs for interpreting CLIP.
For example, the highlights for ``holding'' focus around the person's hands (the 5th row of Fig.~\ref{fig:vis_comp}i), while ``standing'' highlights the person's legs (the 6th row of Fig.~\ref{fig:vis_comp}i). We also notice that the sticks in the background are highlighted, which is probably because the sticks are regarded as ``standing'' in the snow. 

In contrast to our methods, Grad-CAM and MaskCLIP can produce highlights on the explained object, but both also generate significant noise. CLIPSurgery tends to put high and coarse attention on the object region, but also contains background noise. M2IB and MaskCLIP fail when the texts are verbs (``holding'' and ``standing''), while RISE performs the worst when interpreting the CLIP model. The results of GAME and Rollout, which are both based on the self-attention of the model, generate confusing heat maps due to the sparse attention between tokens in some layers.

\subsubsection{Explanations on image-sentence pairs}
\label{sec:vis_img_sentence}

The explanation map from Grad-ECLIP can also be generated from the text encoder viewpoint. 
Using the gradient of the matching score and the feature embeddings of word tokens, Grad-ECLIP can show the importance of each word in the given sentence when matching with an image. Fig.~\ref{fig:vis_img_txt} shows example visual and textual explanations for image-text pairs from MS COCO.
Although Rollout and GAME can highlight important words in the sentence, Grad-ECLIP is the only one showing good correspondence between image attention regions and important words. From the explanation of the sentence, we can identify which words are more important for CLIP when matching with the specific image, and correspndingly the text-specific important regions on the image are shown in the visual explanation. This word importance visualization of the input text can be helpful when designing text prompts for image-text dual-encoders in practical applications.

\subsubsection{\zcy{Visualization examples on diverse image domains}} \label{sec:vis_domain}

We show the visualization comparison of different methods on the samples from different image domains, including the original ImageNet, ImageNet in different domains: rendition  (
pencil sketch (ImageNet-Sketch \cite{wang2019learning}), natural adversarial example (ImageNet-A \cite{hendrycks2021natural}), web images with captions (Conceptual Captions (CC) \cite{sharma2018conceptual}), and chest x-ray with text (MSCXR \cite{boecking2022making}) in Fig.~\ref{fig:vis_domains}. For the image-caption pairs from the web-collected CC and chest X-ray data MSCXR, we generate explanations for both the image and text encoders, and compare with the other methods that also provide text encoder explanations, including the raw attention, Rollout, GAME, and M2IB. 

Our Grad-ECLIP explanations provide interesting insights into how CLIP handles different image domains. In Fig.~\ref{fig:vis_domains} (top), given a normal banana image and text ``banana'', Grad-ECLIP reveals that the yellow color is dominant to CLIP. However, when given a pencil sketch without color (ImageNet-Sketch), Grad-ECLIP reveals that CLIP looks at the curvature of the banana. For the color sketch of the banana (ImageNet-R), Grad-ECLIP shows that the color of the banana is mainly used, and not the black curved lines. Thus, from these examples, we may infer that CLIP prefers using the yellow color over the curved shape for matching with the ``banana'' text. 
%
%
With the sample of a web image and caption in the CC dataset, Grad-ECLIP generates \red{visual highlights}
and the corresponding textual explanation map, showing that ``cowboy'', ``horses'' and ``desert'' are the main concepts used in the matching (from high to low importance).

Grad-ECLIP also provides interesting insights into how the original CLIP fails on novel domains.   The last 2 rows of Fig.~\ref{fig:vis_domains} show the explanations for chest x-ray images and text for the OpenAI CLIP model and a fine-tuned CLIP model (on MSCXR).   The Grad-ECLIP explanation shows that the original CLIP uses the whole lobe to match with the words ``defined'' and ``lobe''. In contrast, the fine-tuned CLIP locates the actual anomaly and matches it with the text ``defined opacities largely''. The reason is that the fine-tuned model is trained to the specific domain that matches the X-ray and the illness location descriptions, while the original OpenAI CLIP model is more general, and apparently ``lobe'' is the keyword and the main object in the image-text pair.

\begin{figure}[t]
\centering
\begin{center}
\includegraphics[width=0.4\textwidth]{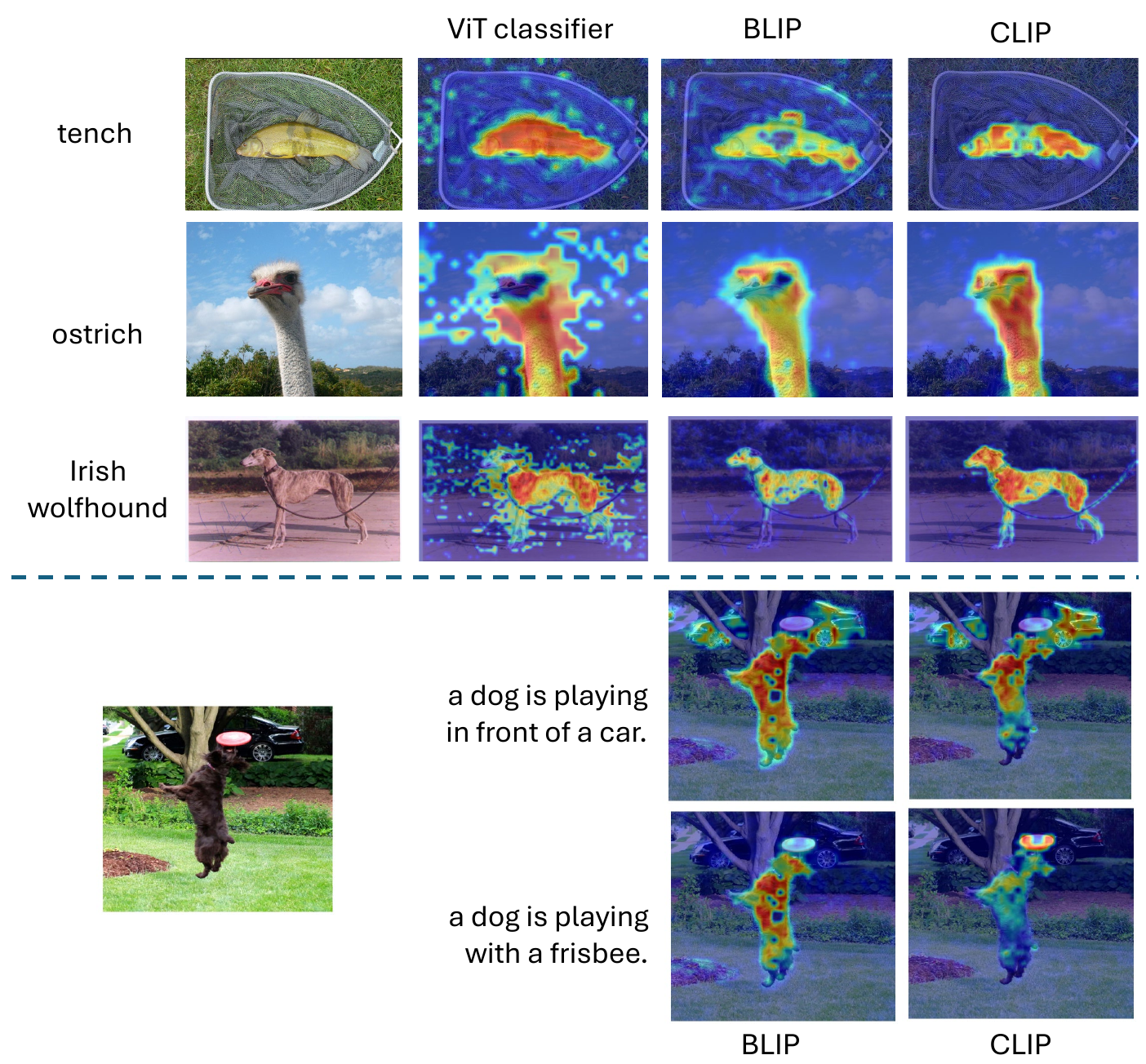}
\end{center}
\vspace{-0.3cm}
\caption{Grad-ECLIP visual explanations of the ViT classifier and BLIP.}
\label{fig:vis_adapt}
\vspace{-0.5cm}
\end{figure}

\CUT{
\begin{table}[t]
\captionof{table}{Evaluation of Grad-ECLIP on the deletion/insertion perturbation metric for ViT-based classifier.}
\label{tab:exp_vit}
\vspace{-0.1cm}
\centering
\scriptsize
\setlength\tabcolsep{1.0pt}    
\renewcommand\arraystretch{1.2}
\begin{tabular}{@{}l|cccccc@{}}
\hline
Method & Grad-CAM\cite{selvaraju2017grad} & ViT-cx\cite{xie2022vit} & T-Attr\cite{chefer2021transformer} & Bi-Attn\cite{chen2022beyond} & TIS\cite{englebert2023explaining} & Grad-ECLIP(ours)\\  \hline
Deletion$\downarrow$  & 0.241 & 0.236 & 0.232 & 0.218 & 0.196  & \textbf{0.174}\\       
Insertion$\uparrow$ & 0.737 & 0.722 & 0.741 & 0.760 & 0.761 & 0.727 \\   
\hline
\end{tabular}
\vspace{-0.1cm}	
\end{table}
}

\subsubsection{\zcy{Explaining ViT classifier and BLIP with Grad-ECLIP}} \label{sec:adapt_vlms}
\vspace{-0.1cm}

Since our explanation method is designed for CLIP encoders, which are Transformer-based, our method can be easily adapted to generate visual explanations for other Transformer-based models.
Here we adapt Grad-ECLIP to a ViT-based classifier \cite{dosovitskiy2020image} and BLIP \cite{li2022blip} to show the generality of our method.

For explaining the ViT-based classifier, the classification score on the corresponding category is used to calculate the gradient and generate the heat map. From the examples for the ViT classifier in Fig.~\ref{fig:vis_adapt}, we can see that although there is some noise on the background, Grad-ECLIP can well mark out the important region on the image for the specific class. 
\CUT{We also conduct the perturbation experiment with the code provided by the repository for deletion/insertion perturbation metric \cite{petsiuk2018rise} and compare with other explanation method for ViT-based classifier. The definition of deletion and insertion metric can be found in the quantitative evaluation section \S\ref{sec:del_ins}. Concretely, in this experiment, ViT-base 16x16 model is used and all input images are 224x224px, with performing a total of 224 perturbation steps and a step size of 224px. Then, the classification score changes are recorded along with the steps for calculating AUC (area under curve) as the deletion or insertion result. The results in Tab.~\ref{tab:exp_vit} compare with other related ViT explanation methods over 5000 ImageNet images with 5 images per class.
\zcyNOTE{Here we don't know how they sample the 5 images in one class. I use range(0,50,10) function. So, we may use the different images.}
From the quantitative evaluation, when adapting Grad-ECLIP to ViT-classifier, our method have an excellent performance with the best deletion result and a comparable insertion result, which indicates that the high responses on the heat map can successfully cover the important region for classifying the image.}
As for the VLMs, shown by the visual explanation results presented in Fig.~\ref{fig:vis_adapt}, when matching the same image-text pair, different models put attention on different regions. For example, BLIP notes the fins, while CLIP notes the fish body to match the image to ``tench''. When matching with the sentence ``a dog is playing with a frisbee'', BLIP puts more importance on the dog in the image, while CLIP places more importance on the frisbee.

Other VLMs like ALBEF \cite{li2021align} add additional attention layers after the encoders to fuse the image and text features, and thus our current method is not directly applicable since our method assumes that the last layer attention output has a linear relationship with the final feature embedding. Our future work will investigate adapting our method to these modified ViT frameworks, e.g., the ALBEF model that uses cross attention to fuse image and text. Nonetheless, our ability to explain CLIP and other VLMs with similar architecture is significant considering that CLIP is by far the most widely used VLM.

\begin{figure}[t]
\centering
\begin{center}
\includegraphics[width=0.4\textwidth]{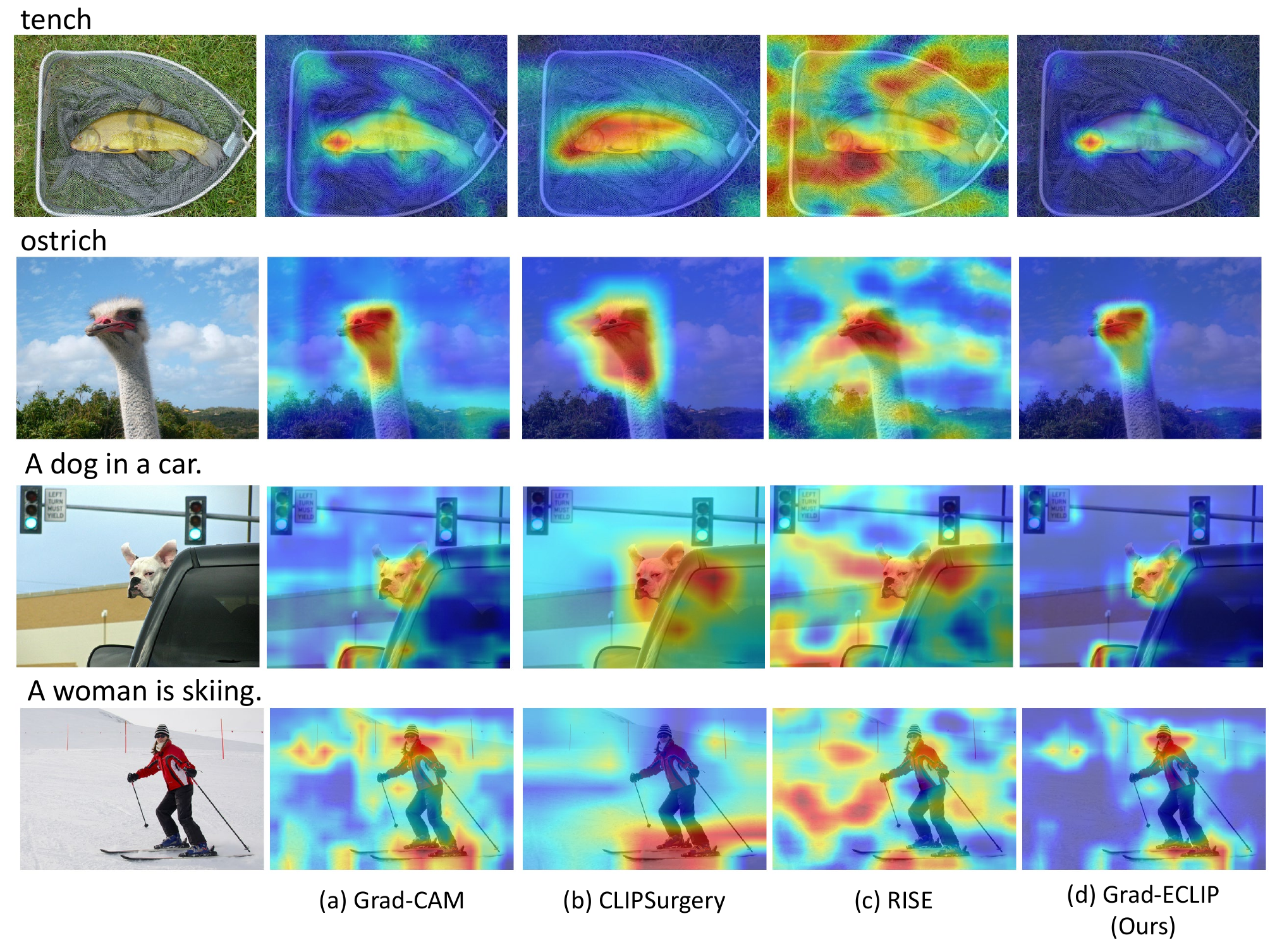}
\end{center}
\vspace{-0.3cm}
\caption{Visual explanations of the CNN-based CLIP with ResNet50 architecture using Grad-ECLIP and other visual explanation methods.}
\label{fig:vis_cnn}
\vspace{-0.5cm}
\end{figure}

\subsubsection{\zcy{Applying Grad-ECLIP to explain CNN-based CLIP}} \label{sec:apply_cnn_clip}

Although the methodology in \S\ref{sec:grad-eclip} for Grad-ECLIP is derived from  CLIP with Transformer architecture, here we show that our method is also applicable to CNN-based CLIP by using the attention layer in the final attention pooling. Figure~\ref{fig:vis_cnn} shows the visual explanation results for ResNet50-CLIP using Grad-ECLIP, and compared to other explanation methods that are compatible with CNN-based CLIP, including Grad-CAM, CLIPSurgery, and RISE. 
%
From the examples, our Grad-ECLIP is able to generate 
text-specific heat maps	\red{with less noise} for interpreting CNN-based CLIP.
In contrast to our method, Grad-CAM can produce similar highlights on the explained objects, but its heat map is noisier in the background. Similar to its performance on ViT-based CLIP, CLIPSurgery generates rougher heat maps, which tend to put high values on a coarse region of the object. Meanwhile, RISE fails to explain the CLIP model.

\begin{figure*}[thp]
	\centering
	\begin{center}
		\includegraphics[width=0.9\textwidth]{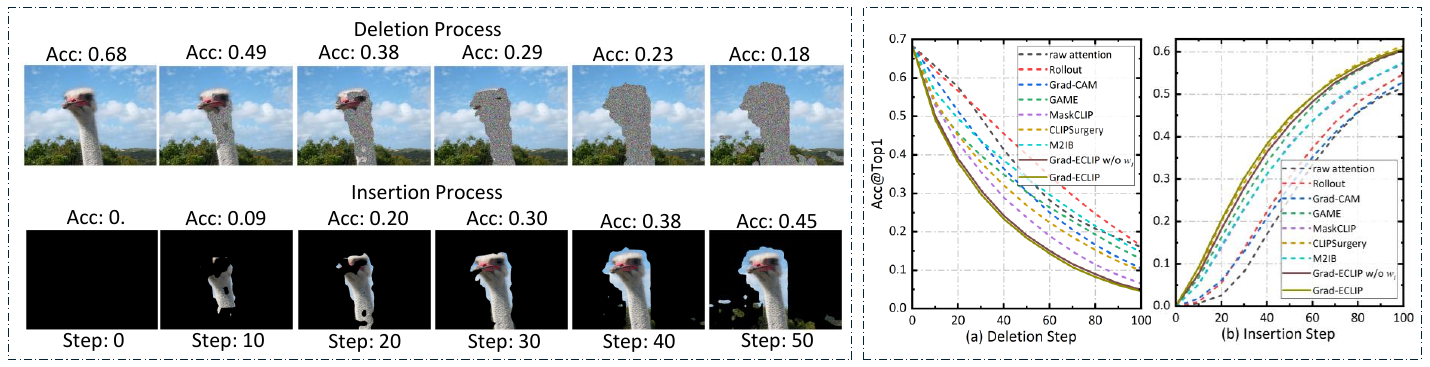}
	\end{center}
	\vspace{-0.3cm}
	\caption{\red{(left) An example showing the deletion and insertion process according to the Grad-ECLIP heat map on an image for 0 to 50 steps. The classification accuracy on each step is shown on the top, which forms the points of the deletion or insertion curves.
	(right)} Classification accuracy at Top-1 vs. (a) Deletion steps and (b) Insertion steps, on the ImageNet validation dataset with visual explanation heat maps from our Grad-ECLIP (solid) and other methods (dashed).
	}
	\label{fig:faithfulness}
\end{figure*}

\subsection{Quantitative evaluation}
\label{exp:quan_eval}
\vspace{-0.1cm}
In this section, we perform quantitative evaluations of Grad-ECLIP compared with baselines.  
In \S\ref{sec:del_ins}, 
the explanation faithfulness is evaluated by the Deletion and Insertion metrics \cite{samek2016evaluating,chattopadhay2018grad,wang2020score,wang2020ss,petsiuk2021black}, which are also called perturbation tests \cite{chefer2021transformer, chefer2021generic}. 
Moreover, in \S\ref{sec:pg_segtest},  we evaluate the localization ability of CLIP using our Grad-ECLIP and the compared methods, when considering each visualization as a soft-segmentation of the image, using PointGame \cite{zhang2018top, chenyang2022odam} and segmentation tests \cite{chefer2021transformer}. \red{Then, we conduct a user trust study to evaluate the trustability of the visual explanations and report the results in \S\ref{sec:user_study}.}
Finally, in \S\ref{sec:process_time}, we evaluate the processing time of Grad-ECLIP compared with other visual explanation methods. 
In this section, in order to understand the effect of the spatial importance term in (\ref{eq:hm}), we also present Grad-ECLIP without using the spatial weight $\lambda_i$, i.e., setting $\lambda_i=1$, which is denoted as 
``w/o $\lambda_{i}$''.


\subsubsection{Deletion and Insertion}\label{sec:del_ins}

\red{The insertion/deletion process for an example image over the first 50 steps is displayed in Fig.~8 left.  To summarize the insertion and deletion curves, we compute the area under the curve (AUC), 
	\begin{align}
		\mathrm{Deletion} = \frac{1}{N_{step}}\sum_{n=1}^{N_{step}} Acc_{del}^{(n)}, \\ \quad
		\mathrm{Insertion} = \frac{1}{N_{step}}\sum_{n=1}^{N_{step}} Acc_{ins}^{(n)},
	\end{align}
	where $Acc_{ins}^{(n)}$ and $Acc_{del}^{(n)}$ are the model performance (e.g., accuracy) with inserted and deleted images at step $n$.  For the deletion curve, a more faithful model will have larger drops in model performance (e.g., accuracy) as pixels are removed, and thus lower AUC for deletion curves is better. On the opposite, for the insertion curve, a more faithful model will have a larger increase in model performance as pixels are added, and thus a higher AUC for insertion curves is better.  We also define an overall metric IMD (Insertion Minus Deletion), which summarizes the two AUCs:
	\begin{align}
		\mathrm{IMD} =\dfrac{1}{N_{step}}\sum_{n=1}^{N_{step}} Acc_{ins}^{(n)}-Acc_{del}^{(n)}.
	\end{align}
	Higher IMD indicates better faithfulness, corresponding to higher insertion AUC and lower deletion AUC.}
We consider each step as 0.5\% of the number of image pixels, and record results for $100$ steps. The model performance is measured using top-1 or top-5 zero-shot classification accuracy on the validation set of %
ImageNet \cite{russakovsky2015imagenet} (ILSVRC) 2012, consisting of 50K images from 1000 classes.
\abcn{In particular, the perturbed image and each of the class names is fed into CLIP, and then classes with the highest scores are selected.}

The insertion/deletion curves for top-1 accuracy are presented in Fig.~\ref{fig:faithfulness} \red{right}, and AUC values for top-1 and top-5 accuracy are presented in Tab.~\ref{tab:del_ins_imagenet}.
Our method obtains the fastest performance drop for Deletion and the largest performance increase for Insertion compared with most related works \red{(p$<$.001)}, showing that regions highlighted in our heat maps better represent explanations of CLIP. CLIPSurgery has comparable results to ours for Insertion \red{(p$=$0.374, 0.413, 0.074, 0.425)}, while performing poorly when evaluated with Deletion. The reason is that CLIPSurgery exhibits heat maps with nearly the same high values 
on the explained target region, so that the deletion operation fails to delete the most important pixels on the image at the beginning steps, which causes the deletion curve to decrease gradually, producing the high deletion AUC. Since CLIPSurgery can locate the explanation target with high values on the heat map, it performs well in the Insertion test. \red{Overall, combining the deletion and insertion, our method has achieved the highest results in terms of IMD (p$<$.001), which shows that changing image inputs based on Grad-ECLIP maps produces the greatest impact on the model prediction.} Our method without using the 
spatial importance (w/o $\lambda_{i}$) has slightly worse performance, but is still better than other baselines. 
As with \cite{chefer2021transformer, chefer2021generic}, we also use both the ground-truth class and the predicted class as the text prompt to generate heat maps, and our method is consistent with them, showing gains when using ground-truth text prompts.

\begin{table*}[tph]
	\vspace{-0.1cm}	
	\captionof{table}{Faithfulness evaluation of \textbf{image} explanation on the \textit{ImageNet} validation set: \red{Deletion, Insertion and IMD (Insertion minus Deletion)}, based on  Top-1 (@1) or Top-5 (@5) classification accuracy. Either the ground-truth or the prediction is used as the text input into CLIP. \red{The arrows indicate lower/higher value is better. The last row shows the p-value for a paired sample t-test between the top-2 methods (bolded and underlined), excluding Ours w/o $\lambda_i$. Significant values (p$<$.05) are highlighted in blue.
	}}
	\vspace{-0.1cm}
	\label{tab:del_ins_imagenet}
	\centering
	\footnotesize
	\setlength\tabcolsep{4.0pt}    
	\renewcommand\arraystretch{1.2}  
	\begin{tabular}{@{}l|cc|cc|cc|cc|cc|cc@{}}
		\hline
		
		& \multicolumn{4}{c|}{Deletion$\downarrow$} & \multicolumn{4}{c|}{Insertion$\uparrow$} & \multicolumn{4}{c}{\red{IMD $\uparrow$}} \\ 
		& \multicolumn{2}{c|}{Ground-truth} & \multicolumn{2}{c|}{Prediction} & \multicolumn{2}{c|}{Ground-truth} & \multicolumn{2}{c|}{Prediction} & \multicolumn{2}{c|}{Ground-truth} & \multicolumn{2}{c}{Prediction}\\
		Method & @1 & @5 & @1 & @5  & @1 & @5 & @1 & @5  & @1 & @5 & @1 & @5\\ 
		\hline
		raw attention  & 0.383 & 0.624  & - & - & 0.249 & 0.419 & -  & - & -0.134 & -0.204 & - & - \\ 
		Rollout & 0.408 & 0.656 & - & - & 0.280 & 0.467 & - & - & -0.128 & -0.189 & - & - \\  
		Grad-CAM  & 0.342 & 0.563 & 0.352 & 0.582 & 0.268 & 0.445 & 0.253 & 0.421 & -0.074 & -0.117 & -0.099 & -0.161 \\ 
		GAME  & 0.336 & 0.573 & 0.350 & 0.594 & 0.361 & 0.564 & 0.343 & 0.538 & 0.025 & -0.010 & -0.007 & -0.055 \\ 
		MaskCLIP  & \underline{0.285} & \underline{0.489} & \underline{0.289} & \underline{0.496} & 0.334 & 0.535 & 0.328 & 0.527 & 0.049 & 0.047 & 0.039 & \underline{0.031} \\
		CLIPSurgery & 0.312 &	0.524 & 0.322 & 0.541 &	\underline{0.383}	& \textbf{0.602} &	\textbf{0.373} & \underline{0.572} & \underline{0.070} & \underline{0.079} & \underline{0.051} & 0.030 \\
		M2IB & 0.363 &	0.595 & 0.363	& 0.595 & 0.335 &	0.541 & 0.335 & 0.541 & -0.028 & -0.054 & -0.029 & -0.054 \\
		Ours  & \textbf{0.246} & \textbf{0.427} & \textbf{0.254} & \textbf{0.442}  & \textbf{0.384} & \underline{0.599} & \underline{0.367} & \textbf{0.575} & \textbf{0.137} & \textbf{0.172} & \textbf{0.113} & \textbf{0.133} \\ 
		
		\hspace{10pt} - w/o $\lambda_{i}$ & 0.254 & 0.438 & 0.263 & 0.457 & 0.372 & 0.583 &  0.353 & 0.556 & 0.118 & 0.145 & 0.089 & 0.099 \\ \hline
		\red{p-value} & \bblue{$<$.001}  & \bblue{$<$.001} & \bblue{$<$.001}	& \bblue{$<$.001} & 0.374 &	0.413 & 0.074 & 0.425 & \bblue{$<$.001} & \bblue{$<$.001} & \bblue{$<$.001} & \bblue{$<$.001} \\
		\hline
	\end{tabular}
\end{table*}
\vspace{-0.1cm}	

\begin{table*}[tph]
	\vspace{-0.1cm}	
	\captionof{table}{Evaluation of \textbf{image} explanation faithfulness on \textit{MS COCO image-text retrieval (Karpathy's split)} validation dataset: \red{Deletion, Insertion and IMD} with performance on image retrieval (IR) and text retrieval (TR) tasks. \red{The last row shows the p-value for a paired sample t-test between the top-2 methods (bolded and underlined), excluding Ours w/o $\lambda_i$. Significant values (p$<$.05) are highlighted in blue.}}
	\label{tab:del_ins_mscoco}
	\centering
	\footnotesize
	\setlength\tabcolsep{4.0pt}  
	\renewcommand\arraystretch{1.2}  
	\vspace{-0.1cm}	
	\begin{tabular}{@{}l|cc|cc|cc|cc|cc|cc@{}}
		\hline
		
		& \multicolumn{4}{c|}{Deletion$\downarrow$} & \multicolumn{4}{c|}{Insertion$\uparrow$} & \multicolumn{4}{c}{\red{IMD$\uparrow$}} \\ 
		& \multicolumn{2}{c|}{IR} & \multicolumn{2}{c|}{TR} & \multicolumn{2}{c|}{IR} & \multicolumn{2}{c|}{TR} & \multicolumn{2}{c|}{IR} & \multicolumn{2}{c}{TR} \\ 
		Method & @1 & @5 & @1 & @5 & @1 & @5 &  @1 & @5  & @1 & @5  & @1 & @5  \\ \hline
		raw attention  & 0.171 & 0.355 & 0.192 & 0.372  & 0.125 & 0.255 & 0.154 & 0.297 & -0.046 & -0.038 & -0.100 & -0.075 \\       
		Rollout   & 0.195 & 0.395 & 0.227 & 0.424  & 0.129 & 0.293 & 0.175 & 0.350 & -0.065 & -0.052 & -0.128 & -0.105 \\ 
		Grad-CAM   & 0.172 & 0.350 & 0.216 & 0.401 & 0.103 & 0.222 & 0.115 & 0.233 & -0.069 & -0.101 & -0.129 & -0.168 \\
		GAME  & 0.171 & 0.355 & 0.198 & 0.380 & \underline{0.154} & \underline{0.308} & \textbf{0.210} & \underline{0.374} & -0.017 & 0.012 & -0.047 & -0.006 \\  
		MaskCLIP   & \underline{0.132} & \underline{0.284} & \textbf{0.152} & \underline{0.295}  & 0.142 & 0.295  & 0.189 & 0.351 & \underline{0.010} & \underline{0.038} & \underline{0.011} & \underline{0.057} \\ 
		CLIPSurgery & 0.179 & 0.365 & 0.238 & 0.429 & 0.142 & 0.294 & 0.177 & 0.338 & -0.037 & -0.061 & -0.071 & -0.091 \\
		M2IB & 0.179 & 0.367 & 0.206 & 0.391 & 0.147 & 0.300 & 0.206 & 0.369 & -0.033 & 0.000 & -0.067 & -0.021 \\
		Ours    & \textbf{0.125} & \textbf{0.267} & \underline{0.155} & \textbf{0.293}  & \textbf{0.158} & \textbf{0.320}  & \underline{0.206} & \textbf{0.376} & \textbf{0.033} & \textbf{0.051} & \textbf{0.053} & \textbf{0.083} \\
		\hspace{10pt} - w/o $\lambda_{i}$ & 0.139  & 0.294  & 0.183 & 0.339 & 0.140 & 0.289 & 0.174 & 0.328  & 0.001 & -0.009 & -0.005 & -0.011 \\  
		\hline
		\red{p-value} & \bblue{$<$.001}  & \bblue{$<$.001} & 0.106 & 0.705 & \bblue{0.011} & \bblue{0.001} & 0.297  & 0.645 & \bblue{$<$.001} & \bblue{$<$.001} & \bblue{0.001} & \bblue{0.001} \\
		\hline
	\end{tabular}
    \vspace{-0.2cm}
\end{table*}

\begin{table}[h]
	\captionof{table}{Evaluation of \textbf{text} explanation faithfulness on \textit{MS COCO image-text retrieval (Karpathy's split)} validation dataset: \red{Deletion, Insertion and IMD} with performance on image retrieval (IR) and text retrieval (TR) tasks. \red{The last row shows the p-value for a paired sample t-test between the top-2 methods (bolded and underlined), excluding Ours w/o $\lambda_i$. Significant values (p$<$.05) are highlighted in blue.}}
	\label{tab:del_ins_text_mscoco}
	\vspace{-0.1cm}	
	\centering
	\footnotesize
	\setlength\tabcolsep{4.0pt}    
	\renewcommand\arraystretch{1.2}
	\begin{tabular}{@{}l|cc|cc|cc@{}}
		\hline
		& \multicolumn{2}{c|}{Deletion$\downarrow$} & \multicolumn{2}{c|}{Insertion$\uparrow$} & \multicolumn{2}{c}{\red{IMD$\uparrow$}} \\ 
		Method & IR & TR & IR & TR & IR & TR \\ \hline
		raw attention  & 0.284 & 0.492  & 0.007 & 0.033 & -0.278 & -0.459   \\       
		Rollout   & 0.122 & 0.239 & 0.105 & 0.207 & -0.017 & -0.032 \\ 
		GAME  &\underline{0.108} & \underline{0.208} &  \underline{0.115}  &  \underline{0.230} & \underline{0.006} & \underline{0.022} \\  
		M2IB  & 0.214 &	0.426 &	0.005 &	0.024 & -0.233 & -0.419 \\
		Ours    & \textbf{0.100} & \textbf{0.177} & \textbf{0.129} & \textbf{0.254}  & \textbf{0.076} & \textbf{0.030} \\
		\hspace{10pt} - w/o $\lambda_{i}$ & 0.112  & 0.211 & 0.112 & 0.236  & 0.001 & 0.025 \\   
		\hline
		\red{p-value} & \bblue{0.019} & \bblue{0.017} & \bblue{0.025} & \bblue{0.026} & \bblue{0.019} & \bblue{0.019} \\
		\hline
	\end{tabular}	
    \vspace{-0.2cm}
\end{table}

\abc{We next evaluate the Deletion and Insertion performance for image and text retrieval tasks on Karpathy's validation split of MS COCO.
To evaluate the image explanations,} image pixels are removed or added step-by-step based on the text-specific explanation heat map. With the modified image replacing the original image, we record the image and text retrieval performance (recall @ top-1 and top-5 matching) to draw the deletion and insertion curve.
Then, the AUC results of Deletion and Insertion with text-specific image explanations are reported in Tab.~\ref{tab:del_ins_mscoco}. 
Grad-ECLIP surpasses the other methods on most metrics \red{(p$<$.001)}, which further demonstrates that our method produces high-quality visual explanation, regardless of whether the text is the class categories as in ImageNet or long captions as in MS COCO.

Finally, we evaluate \textit{the faithfulness of our text explanations} using the \textit{text version} of the Deletion and Insertion metric, 
where words are deleted or inserted based on the order of importance in the text heat map. Using images and caption annotations in MS COCO Karpathy's split, we record the image-text retrieval performance for the modified caption, changing with a total of 5 steps, with one word deleted/inserted at a time. 
The results in Tab.~\ref{tab:del_ins_text_mscoco} show that Grad-ECLIP has the highest faithfulness (best deletion, insertion \red{and IMD} scores \red{with p$=$0.019}) compared with the other Transformer explanation methods.


\red{The last row of Tables I, II, and III displays the paired sample t-test results conducted between the AUCs of the top-2 methods in the column, where pairs correspond to the same step on the two curves.}

\subsubsection{Point Game and Segmentation Test}\label{sec:pg_segtest}
\red{Following the pointing game and segmentation test metric adopted in related works \cite{wang2020score, petsiuk2018rise, wang2023visual, jiang2021layercam, zhang2018top, chefer2021transformer}, we evaluate the localization ability of CLIP using our Grad-ECLIP and the compared methods.}  
\red{It should be noted that these localization tests are mainly a rationalization of the CLIP prediction, rather than a faithfulness metric. Indeed, localization or segmentation performance is not necessarily indicative of the faithfulness, in terms of the insertion and deletion metrics, as indicated in Table~\ref{tab:del_ins_imagenet}.}

We adopt the ImageNet-Segmentation (ImageNet-S) \cite{gao2022luss} validation set, which has segment annotations on 12,419 images of 919 categories from ImageNet. Point Game (PG) is a commonly used metric to evaluate the localization correctness of a visual explanation. PG counts a hit score if the location with the largest value in the 
visual explanation
heat map lies within the object region, which can be defined by the class segmentation mask. Then the PG accuracy is measured by averaging over all samples. Since PG only considers the maximum point, but not the spread of the heat map, we also conduct energy-PG \cite{wang2020score}, which calculates the proportion of heat map energy within the ground-truth mask versus the whole map. Similar to the evaluation by \cite{chefer2021transformer, chefer2021generic}, regarding the heat maps as soft-segmentation results, we adopt pixel accuracy (Pixel Acc.), average precision (AP), and averaged mask intersection over union (maskIoU) as additional metrics. 

The results for localization are shown in Tab.~\ref{tab:image_seg}. 
Both versions of Grad-ECLIP significantly outperform other explanation methods on PG and energy-PG \red{($\chi^2$ test on PG hit numbers between Grad-ECLIP and the second best method CLIPSurgery, $\chi^2$=593.3, df=1, p$<$.001)},
which demonstrates that Grad-ECLIP can well reveal that the important pixels for CLIP are inside the object region.
Comparing Grad-ECLIP with and without $\lambda_i$, Grad-ECLIP without $\lambda_{i}$ obtains relatively higher performance on pixel accuracy and maskIoU, since heat maps that contain more high-value pixels within the ground-truth mask have an advantage on these two metrics. In Fig. 9(b,c) of the main paper, using $\lambda_{i}$ reduces the values on the mask while removing the surrounding noise. Due to a similar reason, CLIPSurgery obtains higher pixel accuracy and maskIoU, since it tends to put high heat map values on all the pixels of the object region, and gets a higher score when aggregating the heatmaps inside the object mask in these two evaluations. However, the lower PG, energy-PG, and AP demonstrate that there are more high values generated outside of the object boundary. 

\begin{table}[th]
\captionof{table}{Evaluation of localization ability using the Point Game (PG and energy-PG) and Segmentation test (Pixel Acc., AP and MaskIoU) on the \textit{ImageNet-S} validation dataset.}
\label{tab:image_seg}
\vspace{-0.1cm}
\centering
\footnotesize
\setlength\tabcolsep{3.5pt}    
\renewcommand\arraystretch{1.2}
\begin{tabular}{@{}l|cc|ccc@{}}
\hline
Method & PG & energy-PG & Pixel Acc. & AP & maskIoU  \\  \hline
raw attention  & 0.1219 & 0.1321 & 0.0278 & 0.2877 & 0.0013 \\       
Rollout & 0.1375 & 0.2835 & 0.2524 & 0.3345 & 0.011 \\   
Grad-CAM & 0.1845 & 0.3154 & 0.5457 & 0.4050 & 0.1251 \\
GAME & 0.4706 & 0.4438 & 0.4765 & 0.4072 & 0.089 \\   
MaskCLIP & 0.4041 & 0.1408 & 0.718 & 0.4557 &  0.2481 \\  
CLIPSurgery	& 0.5759 &	0.3983 &	\textbf{0.7546} &	0.4608 &	\textbf{0.3471} \\
M2IB &	0.2640 &	0.3557 &	0.6194 &	0.4003 & 	0.1474 \\
Ours & \textbf{0.8899} & \textbf{0.5997} & 0.7056 & \textbf{0.5662} & 0.2869 \\  
\hspace{10pt} - w/o $\lambda_{i}$ & 0.8356 & 0.4409 & 0.7365 & 0.5163 & 0.3314 \\  
\hline
\end{tabular}
\vspace{-0.2cm}
\end{table}

\subsubsection{\rred{User Study on Improving Understanding}}
\label{sec:user_study}
\CUT{
\red{In this section, we evaluate the trustability of the visual explanations through the following user trust test. Heat maps are generated by  \textit{Grad-CAM} \cite{selvaraju2017grad}, \textit{GAME} \cite{chefer2021generic}, \textit{MaskCLIP} \cite{zhou2022extract}, \textit{CLIPSurgery} \cite{li2023clipsurgery}, M2IB \cite{wang2024visual}, and our Grad-ECLIP for 20 randomly selected image-text pairs from the MS COCO validation set. For each sample, users\footnote{Informed consents were obtained.} are asked to choose one of the heat maps by the question ``Which heat map best helps you to understand where the model focuses to calculate the similarity of the image-text pair?'' We totally collect 360 responses, 18 for each sample. The form of the questionnaire and one sample are presented in the Appendix.}

\red{In the results, our Grad-ECLIP received significantly more votes ($51.1\%$), compared with other explanation methods: Grad-CAM $7.8\%$, GAME $6.7\%$, MaskCLIP $15.6\%$, CLIPSurgery $13.8\%$, M2IB $5.0\%$, $\chi^2$(120, N=360)=382.7, p$<$.001. The significantly higher human trust of Grad-ECLIP demonstrates its superior image-text matching interpretability for CLIP.}
}

	\rred{
    In this section, we evaluate whether the Grad-ECLIP explanations can improve a user's understanding of CLIP's predictions. 
    We adopt a \emph{forward simulation} task \cite{hase-bansal-2020-evaluating}, where the user is given an image and two captions and is asked to judge which caption will receive a higher matching score from the model under two conditions: with or without corresponding visual explanations provided. If the users' judgment accuracy improves when provided with the explanations, then it suggests that the explanations allow the users' to better understand the model's predictions.}

    \rred{
    Ten images and their corresponding captions were randomly selected from the dataset. A questionnaire was created, where each image was displayed with two text caption options, and the participant was asked to select the caption that they think the CLIP model will give a higher matching score. The images were grouped into two sets of five, denoted as Set I and Set II. In Set I, the image and captions were displayed without explanations, and in Set II they were displayed with the corresponding Grad-ECLIP explanations for each image-caption pair. Before each Set, a few examples of image-caption pairs, their CLIP scores, and explanations (Set II only) were presented to the participant in a training phase.    
    Set I was presented before Set II so that the participant's initial understanding of the model was not biased by the explanations. 
    The assignment of images to Set I and Set II was counterbalanced so that a given image appeared the same number of times in Set I and Set II over all participants.
    We collected responses from a total of 34 participants.
    More details about the questionnaires, including sample selection and examples, are presented in the Appendix.}
	
	\rred{
    The results of the user study indicate that the participants' correct judgment of CLIP's higher-matched sentence significantly improved when the Grad-ECLIP explanations were provided, from 67.6\% to 80.0\% accuracy (paired t-test, t(33)=1.904, p=0.032).   
    Thus, the user study demonstrates that the Grad-ECLIP explanations can effectively assist users in understanding the prediction-making mechanism of the CLIP model, highlighting its superior image-text matching interpretability.}

\begin{table*}[!h]
\vspace{-0.1cm}
\captionof{table}{Comparison of the average processing time (on RTX3090 GPU) per image for generating the explanation map.}
\label{tab:proc_time}
\vspace{-0.1cm}
\centering
\footnotesize
\setlength\tabcolsep{3.5pt}    
\renewcommand\arraystretch{1.2}
\begin{tabular}{@{}l|ccccccccc@{}}
\hline
Method & raw attention & Rollout & Grad-CAM & GAME & MaskCLIP & CLIPSurgery & M2IB & RISE & Grad-ECLIP(Ours)  \\  \hline
time (s/img)  & 0.0117 & 0.0298 & 0.0114 & 0.0228 & 0.0117 & 2.9423 & 0.5781 & 6.2376 & 0.0165 \\       
\hline
\end{tabular}
\vspace{-0.2cm}	
\end{table*}

\subsubsection{Processing time comparison}\label{sec:process_time}

In  Tab.~\ref{tab:proc_time}, we show the average processing time per image, which counts the total duration from inputting the image and text into CLIP to obtaining the explanation map. Since the gradient can be easily and quickly obtained through the autograd function of Pytorch, both our method and Grad-CAM take similar processing time as the raw attention and MaskCLIP, which obtain their heat maps from the forward pass of the model and some other minor operations. Note that for gradient-based methods, the backpropagation does not need to go all the way to the input layer, but stops at an intermediate upper layer, and thus, the extra computation required is not much. RISE needs the longest processing time, which is a common drawback of perturbation-based methods.

\vspace{-0.1cm}
\subsection{Ablation study}
\vspace{-0.1cm}

In this section, we conduct ablation studies to illustrate the influence of the proposed loosened spatial weight (\S\ref{sec:spatial_weight}), the number of involved layers (\S\ref{sec:influnce_layers}), and multi attention heads (\S\ref{sec:influence_heads}) in the calculation of Grad-ECLIP.

\subsubsection{\zcy{Effect of the loosened spatial weight}}
\label{sec:spatial_weight}


Here we conduct an ablation study on the $\lambda_{i}$ in Eq.~\ref{eq:weight} to show the effect of the proposed spatial weight. 
\abc{We consider two versions of Grad-ECLIP with modified spatial importance $\lambda_i$: the first version uses the softmax attention rather than 0-1 normalization (denoted as $\lambda_i=\mathrm{softmax}$); the second version removes the spatial importance completely by setting $\lambda_i=1$.}

\begin{figure}
\vspace{-0.1cm}
\centering
\begin{center}
\includegraphics[width=0.4\textwidth]{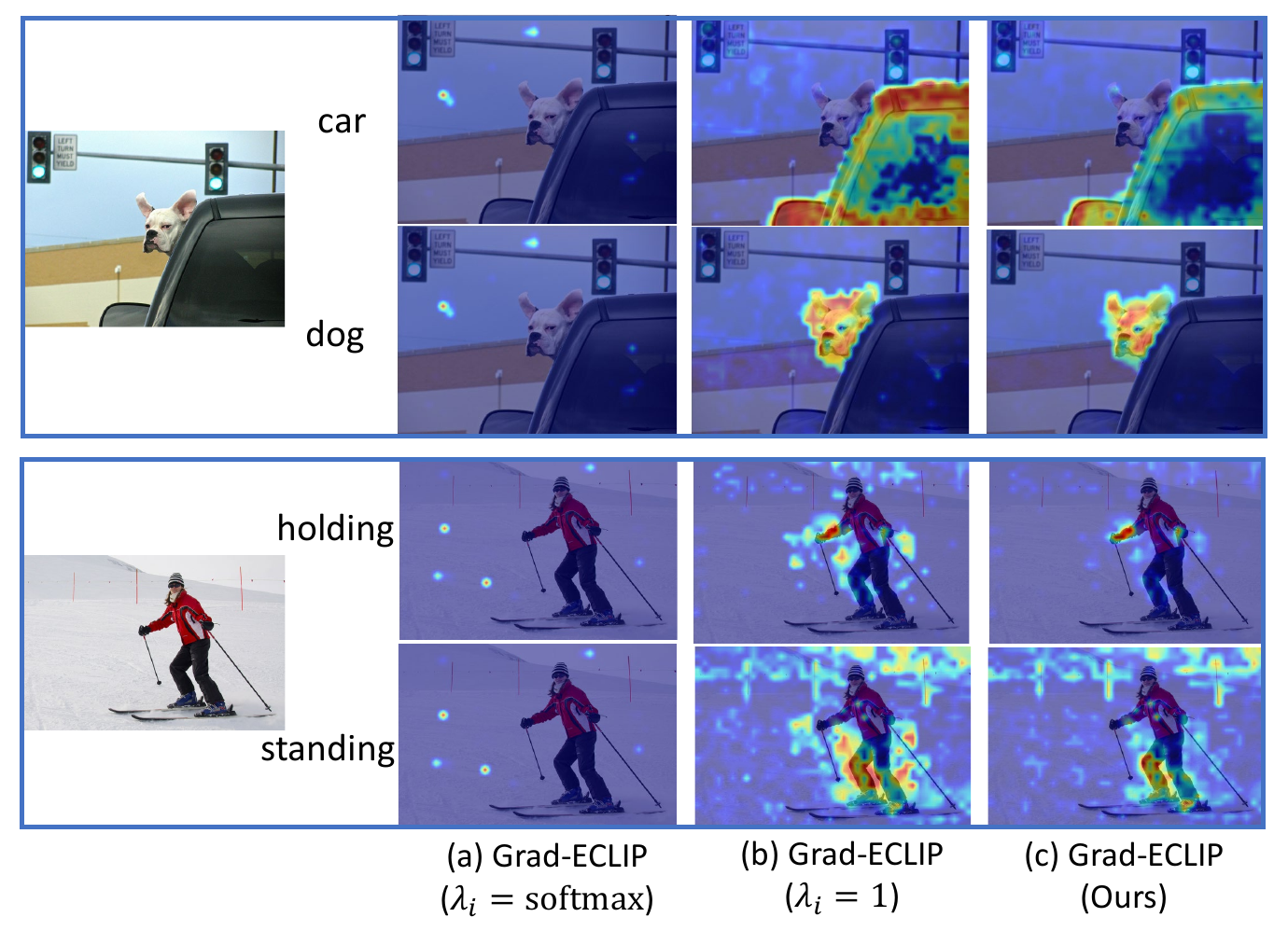}
\end{center}
\vspace{-0.3cm}
\caption{The effect of spatial weight $\lambda_{i}$. We compare the explanation maps of Grad-ECLIP with a version without the proposed $\lambda_i$ (denoted as ``w/o $\lambda_i$'', which replaces the proposed spatial weights with (a) $\lambda_i=\mathrm{softmax}$, (b) $\lambda_i=1$), and (c) the full Grad-ECLIP.}
\vspace{-0.2cm}
\label{fig:w_i}
\end{figure}

A comparison of visualizations is presented in  Fig.~\ref{fig:w_i}.
Compared with the heat maps generated by the full Grad-ECLIP in Fig.~\ref{fig:w_i}c, the version that removes spatial importance altogether ($\lambda_{i}=1$) contains more noise near object boundaries and on the background (Fig.~\ref{fig:w_i}b), but is otherwise consistent with full Grad-ECLIP. The result of using $\lambda_{i}=\mathrm{softmax}$ (Fig.~\ref{fig:w_i}a) is equivalent to raw attention (Fig.~\ref{fig:vis_comp}a) due to the output of the $\mathrm{softmax}$ being extremely sparse. We also provide the quantitative comparisons of Grad-ECLIP using $\lambda_{i}=1$  in the \S\ref{sec:del_ins} and Appendix C, denoted as ``w/o $\lambda_i$''.

\subsubsection{\zcy{Effect of number of layers on visual explanation}}
\label{sec:influnce_layers}
As introduced in \S\ref{sec:grad-eclip}, the explanation can be aggregated over all the layers in Transformer by recursively processing each layer with Eq.~\ref{eq:hm}. In this section, we conduct the experiments to discuss the influence of using different numbers of layers to generate heat maps for images and text.



\red{Let $N_I$ and $N_T$ be the number of layers used for explaining the image and text encoders, respectively, where $N_I=1$ means the visualization was generated with only the final Transformer layer (for image encoder), and $N_I=12$ means that the visualizations are aggregated over all Transformer layers.
	The visualizations for different $N_I$ for image explanations are shown in Fig.~\ref{fig:layers_img}, and the corresponding caption explanations are shown in Fig.~\ref{fig:layers_text} for different $N_T$. }
\red{Table~\ref{tab:del_ins_layer_img} evaluates the image explanation faithfulness vs. the number of transformer layers ($N_I$) aggregated in the image encoder. 
The setting of $N_I=1$ gives the highest IMD.  Figure~\ref{fig:layers_img} shows the corresponding example for different $N_I$ -- the lower layers of the image encoder include less semantic information and may introduce more noise to the heat maps, which is why using only the final layer ($N_I=1$) in the image encoder has higher faithfulness.} Therefore, it is best to just use the last layer in the calculation of image visual explanation, and this conclusion is consistent with the classical gradient-based CAM methods \red{for CNNs}.

As for the text explanation, there is no obvious difference in the visualization quality, since the highlights are basically focusing on “dog”, “car”, and “traffic lights” with some minor variations. \red{The faithfulness evaluation results on the text explanation maps based on different numbers of layers $N_T$ are shown in the following Table~\ref{tab:del_ins_layer_text}.} The explanation faithfulness has the trend that it first increases with more layers used and then goes down with the lower-layer features involved (\red{$N_T > 8$}). Therefore, we aggregate the last eight layers of maps for interpreting the text encoder in our experiments.

\CUT{
\begin{figure}[t]
\vspace{-0.1cm}
\centering
\begin{center}
\includegraphics[width=0.45\textwidth]{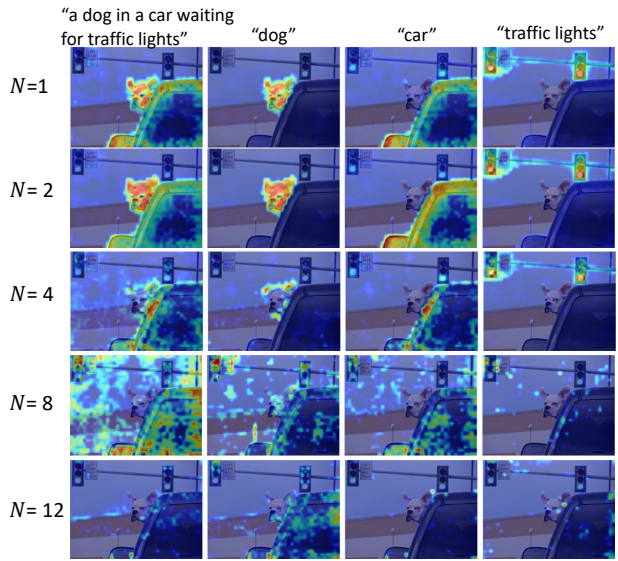}
\end{center}
\vspace{-0.3cm}
\caption{The \textbf{image} visual explanations generated when aggregating over $N$ layers of the image transformer encoder.}
\label{fig:layers_img}
\vspace{-0.1cm}
\end{figure}

\begin{figure}[t]
\vspace{-0.2cm}
\centering
\begin{center}
\includegraphics[width=0.45\textwidth]{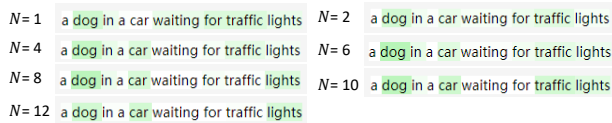}
\end{center}
\vspace{-0.3cm}
\caption{The \textbf{textual} explanations generated when aggregating over  $N$ layers of the text transformer encoder.}
\label{fig:layers_text}
\vspace{-0.5cm}
\end{figure}
}

\begin{figure}[t]
	\vspace{-0.1cm}
	\centering
	\begin{center}
		\includegraphics[width=0.4\textwidth]{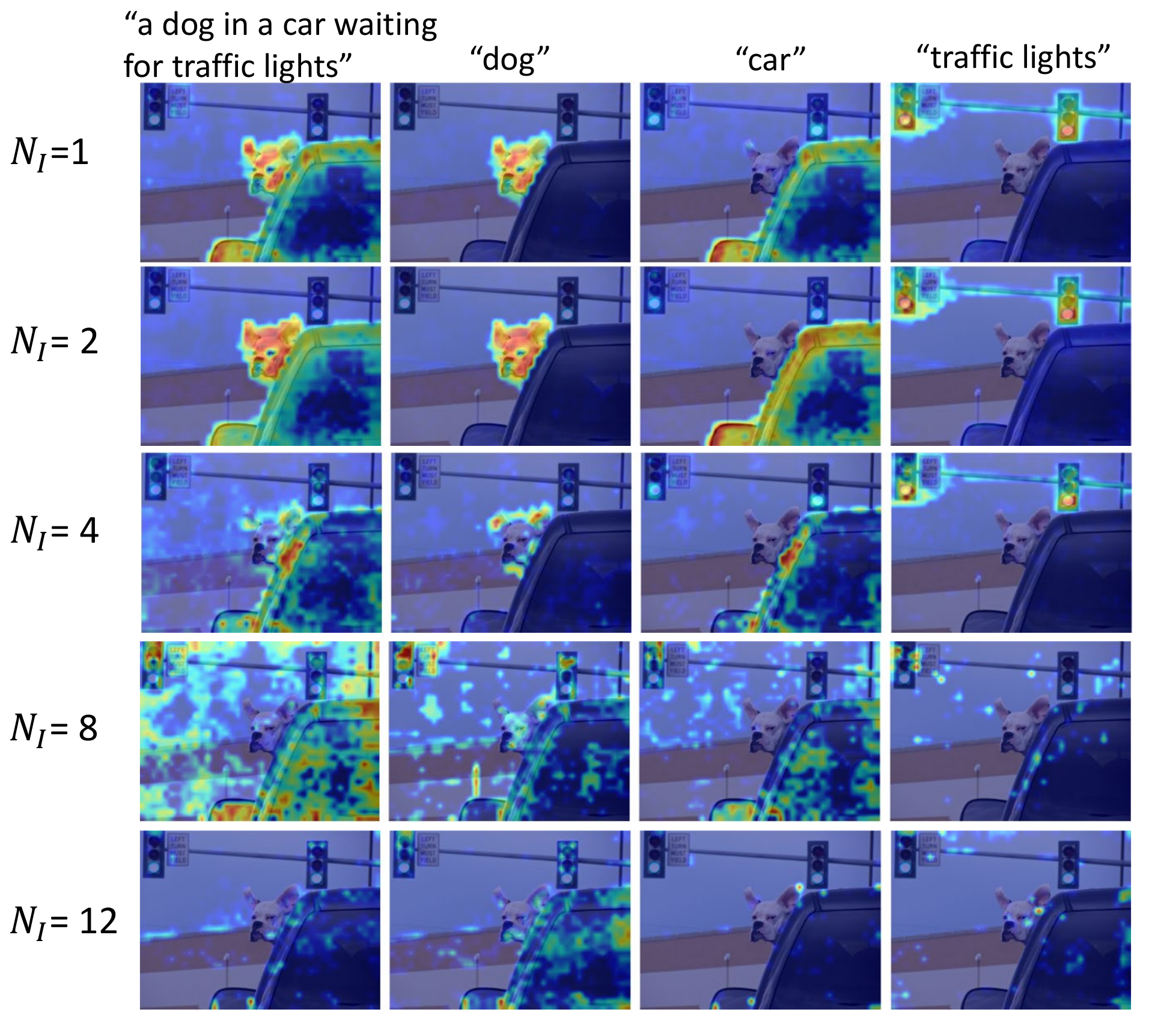}
	\end{center}
	\vspace{-0.3cm}
	\caption{The \textbf{image} visual explanations generated when aggregating over \red{$N_I$} layers of the image transformer encoder.}
	\label{fig:layers_img}
	\vspace{-0.2cm}
\end{figure}

\begin{table}[t]
	\captionof{table}{\red{The \textbf{image} explanation faithfulness vs. the number of transformer layers aggregated for the explanation. Evaluating on \textit{MS COCO image-text retrieval (Karpathy's split)} validation dataset: Deletion, Insertion, and IMD with reporting image retrieval (IR) and text retrieval (TR) performance.}}
	\label{tab:del_ins_layer_img}
	\centering
	\footnotesize
	\setlength
	\tabcolsep{3.5pt}  
	\renewcommand\arraystretch{1.2}  
	\red{
	\begin{tabular}{@{}l|cc|cc|cc@{}}
		\hline
		& \multicolumn{2}{c|}{Deletion$\downarrow$} & \multicolumn{2}{c|}{Insertion$\uparrow$} & \multicolumn{2}{c}{IMD$\uparrow$} \\ 
		\red{$N_I$} & IR & TR & IR & TR & IR & TR \\ \hline
		1  & \textbf{0.1246} & \textbf{0.1551} & \textbf{0.1576} & \textbf{0.2056} & \textbf{0.0330} & \textbf{0.0505} \\       
		2  & 0.1279 & 0.1657 & 0.1547 & 0.2010 & 0.0269 & 0.0253 \\ 
		4  & 0.1622 & 0.2051 & 0.1182 & 0.1399 & -0.0439 & -0.0652 \\   
		8  & 0.2316 & 0.2879 & 0.0484 & 0.0440 & -0.1832 & -0.2439 \\ 
		10 & 0.2121 & 0.2643 & 0.0715 & 0.0774 & -0.1407 & -0.1868 \\ 
		12 & 0.2163 & 0.2642 & 0.0699 & 0.0773 & -0.1465 & -0.1869\\ 
		\hline
	\end{tabular}
	}
	\vspace{-0.1cm}
\end{table} 

\begin{figure}[t]
	\centering
	\begin{center}
		\includegraphics[width=0.45\textwidth]{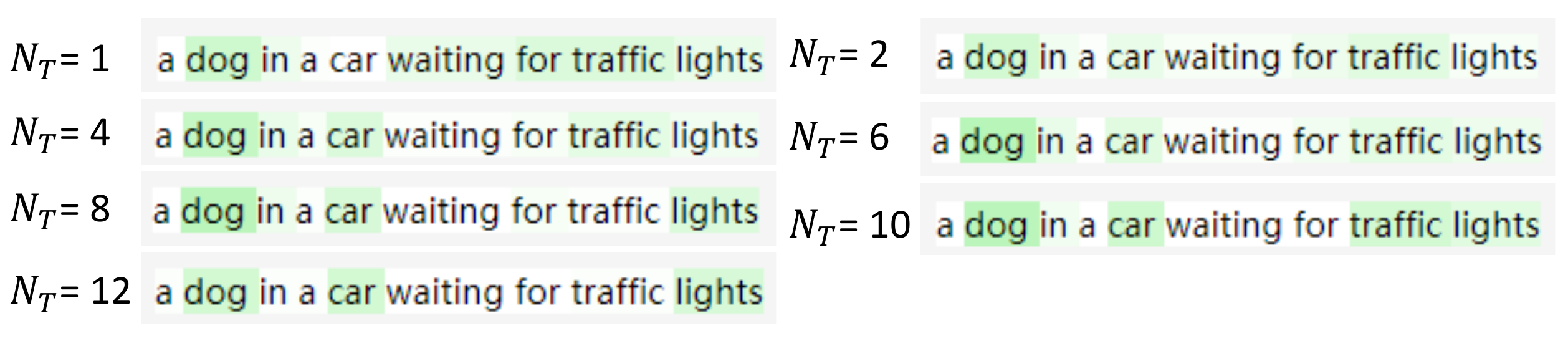}
	\end{center}
	\vspace{-0.3cm}
	\caption{The \textbf{textual} explanations generated when aggregating over \red{$N_T$} layers of the text transformer encoder.}
    \vspace{-0.4cm}
	\label{fig:layers_text}
\end{figure}

\CUT{
\begin{table}
\vspace{-0.3cm}
\captionof{table}{The \textbf{text} explanation faithfulness vs. the number of transformer layers aggregated for the explanation. Evaluating on \textit{MS COCO image-text retrieval (Karpathy's split)} validation dataset: AUC for Deletion and Insertion curves with reporting image retrieval (IR) and text retrieval (TR) performance.}
\vspace{-0.2cm}
\label{tab:del_ins_layer_text}
\centering
\footnotesize
\setlength
\tabcolsep{10.0pt}  
\renewcommand\arraystretch{1.2}  
\begin{tabular}{@{}l|cc|cc@{}}
\hline
& \multicolumn{2}{c|}{Deletion$\downarrow$} & \multicolumn{2}{c}{Insertion$\uparrow$} \\ 
$N$ & IR & TR & IR & TR \\ \hline
1  & 0.1118 & 0.2087 & 0.1059 & 0.2196 \\       
2  & 0.1021 & 0.1826 & 0.1186 & 0.2351 \\ 
4  & \textbf{0.0995} & 0.1786 & 0.1242 & 0.2428 \\  
6  & 0.0989 & \textbf{0.1761} & 0.1273 & 0.2490 \\ 
8  & 0.0996 & 0.1770 & \textbf{0.1292} & \textbf{0.2536} \\ 
10 & 0.1008 & 0.1843 & 0.1288 & 0.2472 \\
12 & 0.1095 & 0.2087 & 0.1219 & 0.2364 \\ 
\hline
\end{tabular}
\vspace{-0.1cm}
\end{table} 
}

\begin{table}[t]
	\captionof{table}{The \textbf{text} explanation faithfulness vs. the number of transformer layers aggregated for the explanation. Evaluating on \textit{MS COCO image-text retrieval (Karpathy's split)} validation dataset: \red{Deletion, Insertion and IMD} with reporting image retrieval (IR) and text retrieval (TR) performance.}
	\label{tab:del_ins_layer_text}
	\centering
	\footnotesize
	\setlength
	\tabcolsep{3.5pt}  
	\renewcommand\arraystretch{1.2}  
	\begin{tabular}{@{}l|cc|cc|cc@{}}
		\hline
		& \multicolumn{2}{c|}{Deletion$\downarrow$} & \multicolumn{2}{c|}{Insertion$\uparrow$} & \multicolumn{2}{c}{\red{IMD$\uparrow$}}\\ 
		\red{$N_T$} & IR & TR & IR & TR & IR & TR \\ \hline
		1  & 0.1118 & 0.2087 & 0.1059 & 0.2196 & -0.0059 & 0.0108 \\       
		2  & 0.1021 & 0.1826 & 0.1186 & 0.2351 & 0.0166 & 0.0525 \\ 
		4  & \textbf{0.0995} & 0.1786 & 0.1242 & 0.2428 & 0.0247 & 0.0643 \\  
		6  & 0.0989 & \textbf{0.1761} & 0.1273 & 0.2490 & 0.0284 & 0.0729 \\ 
		8  & 0.0996 & 0.1770 & \textbf{0.1292} & \textbf{0.2536} & \textbf{0.0296} & \textbf{0.0766} \\ 
		10 & 0.1008 & 0.1843 & 0.1288 & 0.2472 & 0.0279 & 0.0629 \\
		12 & 0.1095 & 0.2087 & 0.1219 & 0.2364 & 0.0125 & 0.0277 \\ 
		\hline
	\end{tabular}
    \vspace{-0.2cm}
\end{table}

\subsubsection{\zcy{Effect of multi attention heads on visual explanation}}
\label{sec:influence_heads}

\begin{figure}[tph]
\vspace{-0.1cm}
\centering
\begin{center}
\includegraphics[width=0.45\textwidth]{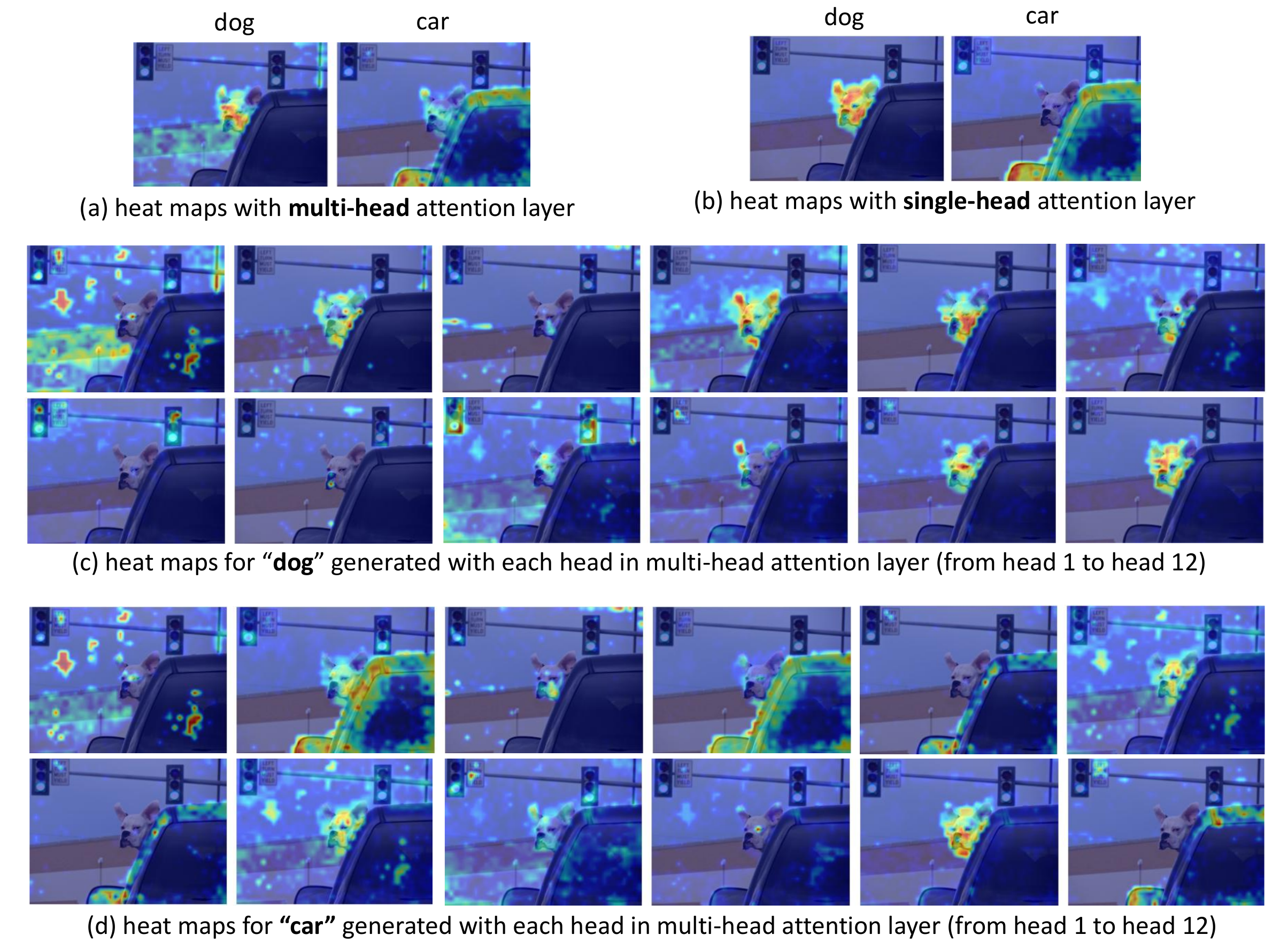}
\end{center}
\vspace{-0.3cm}
\caption{The visual explanation maps using (a) multi-head attention layer; (b) single-head attention layer; (c) each head in the multi-head attention layer for text ``dog''; (d) each head in the multi-head attention layer for text ``car''.}
\vspace{-0.5cm}
\label{fig:heads}
\end{figure}

As mentioned in (\ref{eq:out_cls}) of \S\ref{sec:grad-eclip}, for producing the Grad-ECLIP visual explanation, we set CLIP to perform the forward pass with a single head in the attention layer instead of the original multi-head attention layer. In Fig.~\ref{fig:heads}, we show the visualization of explanation maps when using multi-head attention layers, compared to using a single head. Comparing Fig.~\ref{fig:heads} (a) and (b), using multi-head attention results in some surrounding context information being highlighted with the explained object. 

We further produce the heat maps for \emph{each} attention head, using the $q\in \mathbb{R}^{D/12} $, $k\in \mathbb{R}^{D/12}$, $v\in \mathbb{R}^{D/12}$, and attention output $o_{cls}\in \mathbb{R}^{D/12}$, where $D$ is channel number before going into multi heads, and visualize them in Fig.~\ref{fig:heads} (c) for the target ``dog'', and (d) for the ``car''.
The visual explanation in each head highlights different regions, not only seeing the target object. We can infer that the channels assigned to each head can preserve different information, and the $\mathrm{softmax}$ inside each head helps the model to encode more context information. 
In contrast, with the single head setting, the $\mathrm{softmax}$ is performed over all channels, which selects out the most important information, and our explanation method can show the model's attention on the specific explained target, as shown in Fig.~\ref{fig:heads} (b).

\vspace{-0.1cm}
\section{Analysis of CLIP using Grad-ECLIP}\label{sec:analysis_clip}
\vspace{-0.1cm}

Useful explanation methods can be used to identify failure modes, establish appropriate users' confidence, and give insight to developers to improve models. Therefore, in this section, we use the visual explanation maps \zcy{and textual explanations} generated by Grad-ECLIP to give examples of how to explore the mechanism in text and image matching, and analyze the strengths, weaknesses, and preferences of the CLIP model. We hope that our explanation tool can help researchers discover more interesting properties of pre-trained image-language models and inspire further development of these models.
\abc{Here we analyze three aspects of CLIP: 1) the decomposition and addibility in image-text matching (\S\ref{sec:concept_decomp}); 2) types of attributes that can be identified by CLIP (\S\ref{sec:diagn_attr}); 3) the concreteness/abstractness of words learned by CLIP (\S\ref{sec:att_concreteness}).}
\red{
We note that there is a risk that the heat map explanations may not be accurate when explaining the model, and we can temper this risk by selecting the XAI that has higher faithfulness to the model.  We have shown in Section IV that our method Grad-ECLIP has the highest faithfulness among the related methods, and thus, this provides a justification for using it to perform our interpretation/analysis of CLIP here.  Future XAI methods may improve on faithfulness and may update and refine these interpretations. Nonetheless, this is the nature of scientific progress.}

\begin{figure}
\centering
\begin{center}
\includegraphics[width=0.45\textwidth]{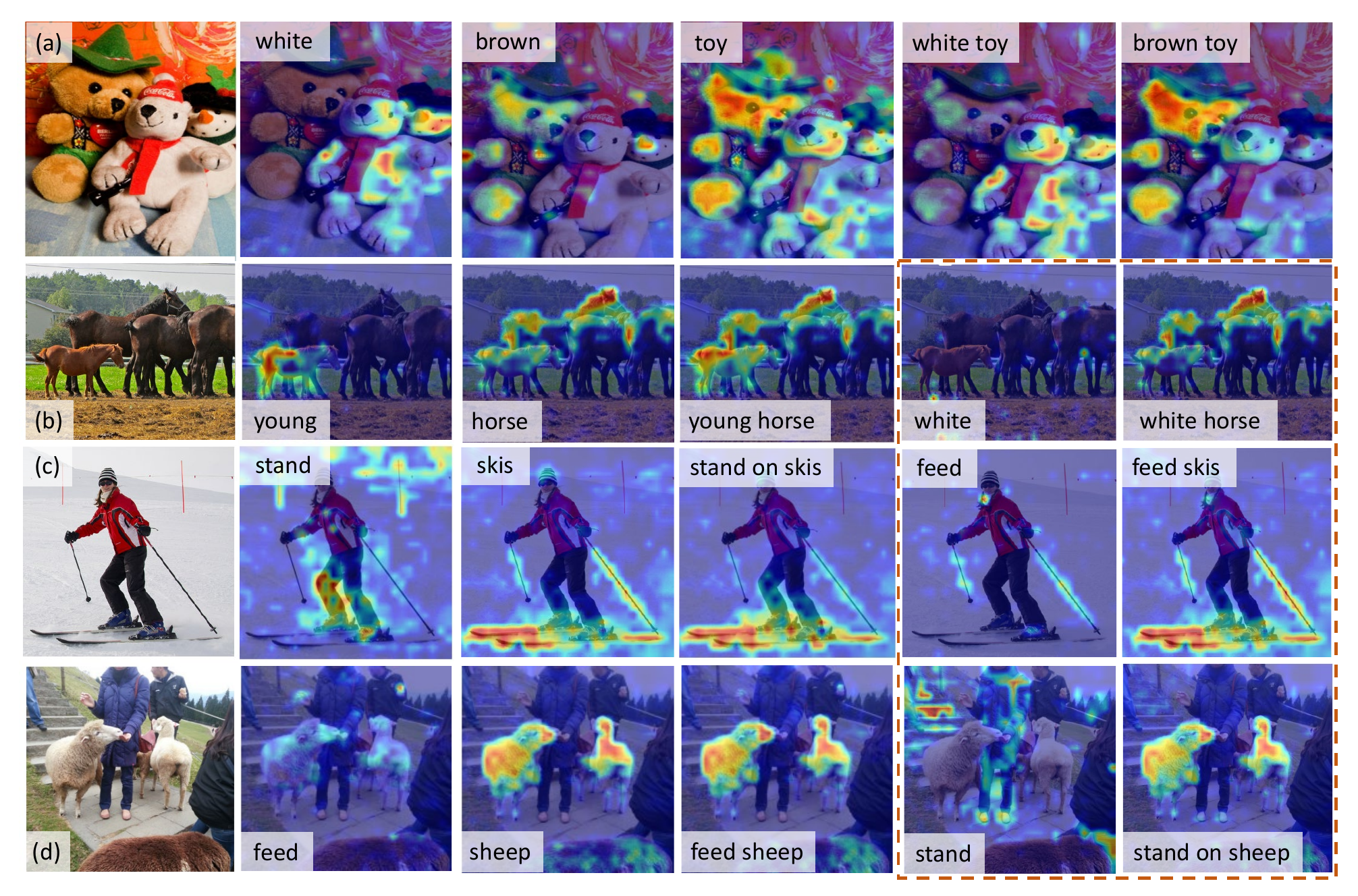}
\end{center}
\vspace{-0.4cm}
\caption{Visual explanation heat maps generated for single words and word phrases using Grad-ECLIP on CLIP. The dashed box contains examples where the text does not match the image.}
\label{fig:word_comb}
\vspace{-0.4cm}
\end{figure}

\vspace{-0.3cm}
\subsection{Concept decomposition and addibility}\label{sec:concept_decomp}
\vspace{-0.1cm}
Examining the visualizations shown in Fig.~\ref{fig:vis_comp}h, CLIP can well recognize the single concepts (nouns) and has good attention to actions (verbs). An interesting question is how it processes the combination of words, \eg, adjective and noun, verb and noun? To examine the working function of phrase matching, we conducted experiments comparing the explanation heat maps for single words and combined phrases using Grad-ECLIP. 

The results are shown in Fig.~\ref{fig:word_comb}. Considering adjective-noun combinations in (a), the highlights are put on all three toys when matching with ``toy'', and CLIP can successfully highlight the correct toy when the color adjective is included in the text. In the case of ``young horse'' in (b), the other horses are still highlighted, while the highlights on the young one are strengthened by adding the attribute ``young''. The examples of verb-noun cases in (c) and (d) also show a similar addibility pattern on the heat maps: (c) with the verb ``stand'', the region of the person's leg is highlighted along with the``skis''; (d) with the verb ``feed'', the people's hands are also highlighted together with sheep. We also show some non-existent concepts or strange word combinations in the dashed box of Fig.~\ref{fig:word_comb}, e.g., ``white horse'' in (b), ``feed skis'' in (c), ``stand on sheep'' in (d).
In these cases, the visualization shows that CLIP will mainly focus on the reasonable part of the concept, such as ``horse'', ``skis'' and ``sheep''. For the non-existent ``white'' concept in image (b), the visual explanation does not highlight anything.

\red{To verify the above observation quantitatively, we conduct an experiment on 60 randomly selected images from the MS COCO validation set. Concretely, for each image, we provide three words based on the content of the image, including one \emph{base word}, one \emph{mismatched word}, and one \emph{matched word}. For example, in Figure.~\ref{fig:word_comb}(b), the base word is ``horse'', and the matched and mismatched words are ``young'' and ``white'', respectively. Combining the base word with the matched word or mismatched word produces the \emph{matched phrase}, e.g. ``young horse'', and the \emph{mismatched phrase} ``white horse'', respectively. We randomly select 30 images 
	for the combination of adjective and noun, and another 30 images for verb and noun combinations.}

\red{To measure concept addibility, we calculate the cosine similarity between the explanation heat map for the whole phrase, and the combination of corresponding word-based heat maps, e.g., the heat map for the matched phrase ``young horse'' and the combination of maps for ``young'' and for ``horse''. The averaged similarities over all images are shown in Tab.~\ref{tab:addibility}
, where we list the results using different combination methods. Combining word-based maps using addition leads to the highest similarity with the phrase-based heat map (p$<$.001, paired t-test between addition and other methods), regardless of whether it is a matched or mismatched phrase. 
Meanwhile, combinations using multiplication have significantly lower similarity compared to addition, which supports the above addibility hypothesis observed from Fig.~\ref{fig:word_comb}.}

\begin{table}[t]
	\captionof{table}{\red{Average cosine similarity between the heat map of a phrase and the combined heat maps of individual words using different combination methods. The last column shows the p-value for a paired sample t-test between ``add maps'' and other methods.}
	}
	\label{tab:addibility}
	\vspace{-0.1cm}
	\centering
	\scriptsize
	\setlength\tabcolsep{1.5pt}    
	\renewcommand\arraystretch{1.2}
	\begin{tabular}{@{}l|cc|cc|c@{}}
		\hline
		
		& \multicolumn{2}{c|}{Matched Phrase} & \multicolumn{2}{c|}{Mismatched Phrase} & p-value\\ 
		Combination method& adj.+noun & verb+noun & adj.+noun & verb+noun & vs. add maps \\ 
		\hline
		add maps  & 0.9750 & 0.9814 & 0.9480 & 0.9755 &  -  \\ 
		multiply maps & 0.7871 & 0.7231 & 0.5810 & 0.5400 & \bblue{$<$.001}  \\  
		maximum maps & 0.9619 & 0.9741 & 0.9319 & 0.9711 & \bblue{$<$.001} \\ 
		\hline
	\end{tabular}
	\vspace{-0.1cm}	
\end{table}

\begin{figure}[th]
	\centering
	\begin{center}
		\includegraphics[width=0.45\textwidth]{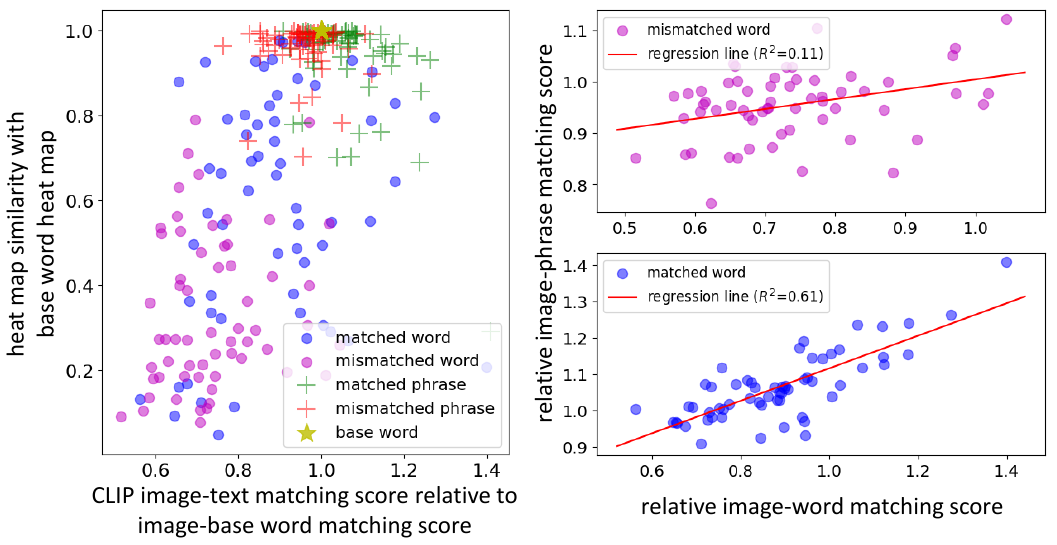}
	\end{center}
    \vspace{-0.2cm}
	\caption{\red{(Left) Scatter plot of explanation heat map similarity vs. CLIP image-text matching score for matched words, mismatched words, matched phrases, and mismatched phrases. The heat map similarity is with the base word, and the CLIP score is relative to the CLIP score using the base word. Thus, all base word points are at coordinates (1,1). (Right) Scatter plot and linear regression of the relative CLIP image-phrase matching score vs. the relative image-word matching score for mismatched words (top-right) and matched words (bottom-right). }}
	\label{fig:decomposition_addibility}
    \vspace{-0.4cm}
\end{figure}

\red{We conducted a further analysis by plotting
	the heat map similarity of the matched/mismatched word or phrase compared to the base word, versus the CLIP image-text matching score, relative to the base word, in Fig.~\ref{fig:decomposition_addibility}(left). 	 
	The mismatched phrase (red ``+'') usually obtains a lower CLIP matching score than the base word (the yellow star) but a higher score than the mismatched word (purple circle), while the explanation heat maps of mismatched phrases have high similarity with the base word (near 1).	
	Therefore, combining with the verified addibility pattern, we infer that when CLIP matches the image with a phrase, it separates the phrase into words and makes the matching result based on putting attention to each word. Thus, for the case of a mismatched phrase (e.g., the white horse in Fig.~\ref{fig:word_comb}(b)) where the object does not exist in the image, the model basically looks at the image region corresponding to the base word ``horse'' and makes the prediction. }
\red{The scatter plot in Fig.~\ref{fig:decomposition_addibility} 
	(top-right) also verifies that the CLIP matching score for mismatched words is usually lower than the base words (lower than 1), while the scores for the corresponding mismatched phrases are close to 1, i.e., similar to just base-word matching score. While there is a linear trend between the relative image-mismatched-word score and the image-phrase score \red{($R^2=0.11$, $p=0.007$ in linear regression), the $R^2$ value is extremely low, indicating significant unexplained variance.}
}
\red{On the other hand, in Fig.~\ref{fig:decomposition_addibility}
	(bottom-right), most matched phrases will have increased score compared to the baseline word, and 
	there is a strong linear trend \red{($R^2=0.61$, $p<.001$)}, where matched words with higher scores will also have higher scores in the matched phrase.}
\red{A matched word can provide further positive effect to using just the base word in the matching process, so that most matched phrases (green ``+'') obtain higher CLIP scores than the base word (the yellow star).}

Therefore, we infer that when processing the matching of the image and phrases, the model has the ability of decomposition and addibility of different concepts. This can help the model to generalize to different scenarios and could be the source of the strong zero-shot ability of CLIP.

\begin{figure}
\vspace{-0.1cm}
\centering
\begin{center}
\includegraphics[width=0.4\textwidth]{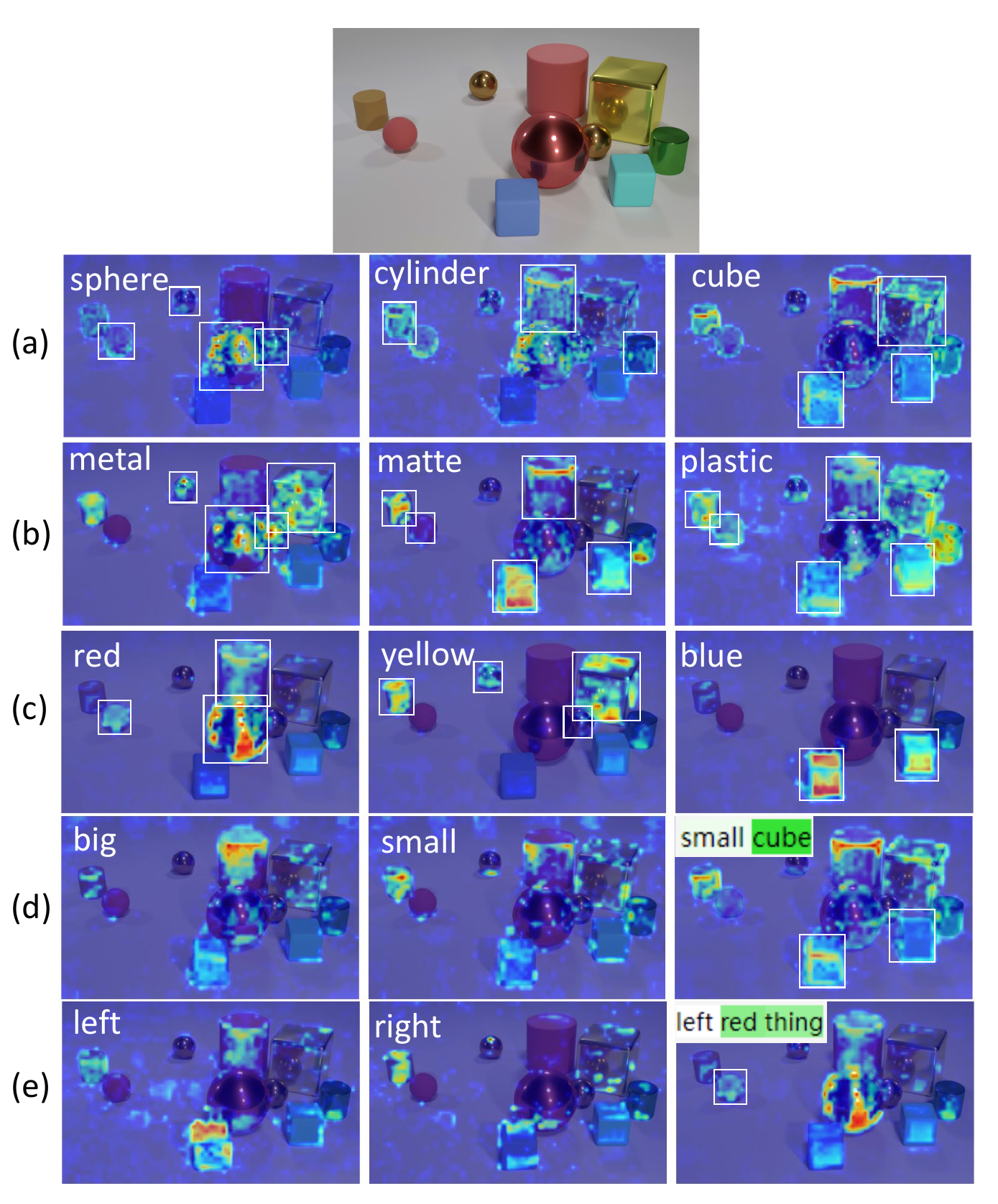}
\end{center}
\vspace{-0.3cm}
\caption{Visual explanations on image matching with different kinds of attributions: (a) shape; (b) material; (c) color; (d) size; (e) position. For visualization, the ground-truth corresponding to the text prompt is outlined with white boxes, except for cases involving relative adjectives, \eg ``big'', ``small'', ``left'', ``right''. The text explanation maps are also shown for ``small cube'' and ``left red thing'' combinations.}
\label{fig:attributes}
\vspace{-0.4cm}
\end{figure}

\vspace{-0.2cm}
\subsection{Diagnostics on attribution identification}\label{sec:diagn_attr}

In Fig.~\ref{fig:word_comb}(a), we see that CLIP has an ability to distinguish color attributes and mark out the corresponding regions on the image. 
To explore further, we conduct an experiment to test CLIP's ability to identify different types of object attributes. We adopt an example image from CLEVR \cite{johnson2017clevr}, a diagnostic dataset for visual reasoning, and visualize image-text matching with various attributes:
shape (sphere, cylinder, cube), material (metal, matte, plastic), color (red, yellow, blue), size (big, small), position (left, right). 

Fig.~\ref{fig:attributes} shows the visual explanation heat maps generated with each image-attribution pair. We have the following findings: 1) for shape and material, the heat maps can show partial correct attention with some obvious objects, such as the metal sphere for ``sphere'' and the highlighted cylinder and cube for ``matte''. However, there are also false positive and false negative errors in (a) and (b). Thus, CLIP possesses a certain but limited knowledge about object shapes and materials. 2) For the color attribute in row (c), the results further verify that the model can have a good ability to distinguish different colors. 3) For comparative attributes, size (big or small) in (d) and position (left or right) in (e), the visual explanations also show that CLIP produces some erroneous results.  For example, there are a few differences between the heat maps of ``small cube'' and ``cube'' in (a), or ``left red thing'' and ``red'' in (c), 
which demonstrates that the words ``cube'' and ``red'' take the major role in the matching. This is also confirmed by the text heat maps in the figure.

\begin{figure}[!h]
	\centering
	\begin{center}
		\includegraphics[width=0.45\textwidth]{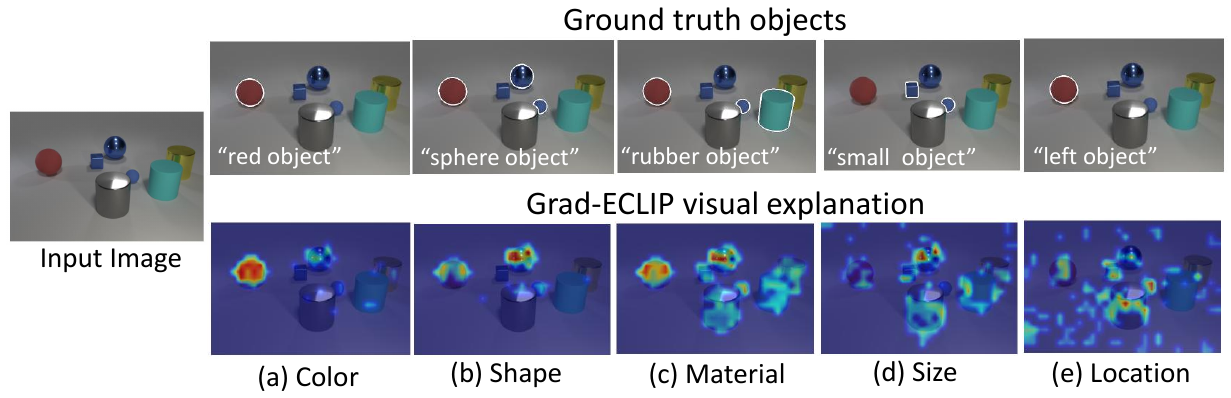}
	\end{center}
    \vspace{-0.1cm}
	\caption{\red{The ground truth 
			objects (outlined in white) and the Grad-ECLIP explanation heat maps corresponding to image-text match for specific attribute words: (a) Color ``red object''; (b) Shape ``sphere object''; (c) Material ``rubber object''; (d) Size ``small object''; (e) Location relationship ``left object''. }
	}
	\label{fig:clver_attribution}
    \vspace{-0.2cm}
\end{figure}

\begin{table}[th]
	\captionof{table}{\red{Averaged similarity/distance between the ground-truth mask and Grad-ECLIP heat maps generated when CLIP matches attribute words in different categories.}}
	\label{tab:clver_attribution}
	\vspace{-0.1cm}
	\centering
	\scriptsize
	\setlength\tabcolsep{3.5pt}    
	\renewcommand\arraystretch{1.2}
	\begin{tabular}{@{}l|ccccc@{}}
		\hline
		& Color & Shape & Material & Size & Location \\ \hline
		
		Cosine similarity $\uparrow$ & \textbf{0.6654} & 0.5959 & 0.5772 & 0.4645 & 0.2758 \\
		Correlation $\uparrow$ & \textbf{0.0419} & 0.0218 & 0.0150  & 0.0100 & 0.0107 \\
		OTD $\downarrow$ & 0.0085 & \textbf{0.0084} & 0.0090 & 0.0093 & 0.0356 \\
		KLD $\downarrow$ & \textbf{2.0237} & 2.0968 & 3.1444 & 4.3012 & 5.0392\\
		JSD $\downarrow$ & \textbf{0.2881} & 0.3222 & 0.3072 & 0.3856 & 0.5327\\
		\hline
	\end{tabular}
	\vspace{-0.1cm}	
\end{table}

\red{We next conduct a quantitative experiment to explore CLIP's ability to encode different object attributes. We utilize the annotations of the CLEVR dataset, which provides the center position of each object with its properties, including color, shape, material, size, and location relationship. With the object coordinates, we use SAM \cite{kirillov2023segany} to generate segmentation masks for the objects, and then perform manual checking to obtain the ground-truth masks.  We randomly select 50 images from the dataset.  
	An image example with the ground truth masks and explanation heat maps for specific attribute text is shown in Fig.~\ref{fig:clver_attribution}.
	Using different attribute words, we measure 
	similarity or distance between the CLIP explanation maps and the corresponding ground-truth, with higher similarity (lower distance) indicating that CLIP encodes relevant features for that attribute by focusing on the corresponding object.
}

\red{
	The results averaged over attribute categories are presented in Tab.~\ref{tab:clver_attribution}.  
	The higher average similarities (or lower distances) for color words demonstrate that the attention of CLIP can better locate the correct object when matching with the given object colors, which means CLIP has a better ability to distinguish or encode color attributes. Following color, the shape and material attributes are the 2nd and 3rd best, while size and location are the worst attributes. 
}
\red{
	To further verify the bad matching with location relationship words, we further conduct an ablation study by comparing using only shape, e.g., ``sphere'', the combination of location and shape, e.g., ``left sphere'', and only location, ``e.g., left'', to describe the left object. The results show that the location relationship words bring no influence to the CLIP matching process, with the heatmap similarity unchanged when adding the location attribute word. See Appendix B for details.}

Overall, from the above analysis, we infer that CLIP has advantages with common perceptual attributes like color, but cannot well handle physical attributes like shape and material, and is weak at grounding objects with comparative attributes, like size and position relationships. Related to the addibility of concepts in the \S\ref{sec:concept_decomp}, 
it is reasonable to expect that attributes that have a concrete visual appearance, e.g., color, will contribute more to the matching score, compared with the abstract comparative attributes.

\begin{figure*}[th]
	\centering
	\begin{center}
		\includegraphics[width=0.85\textwidth]{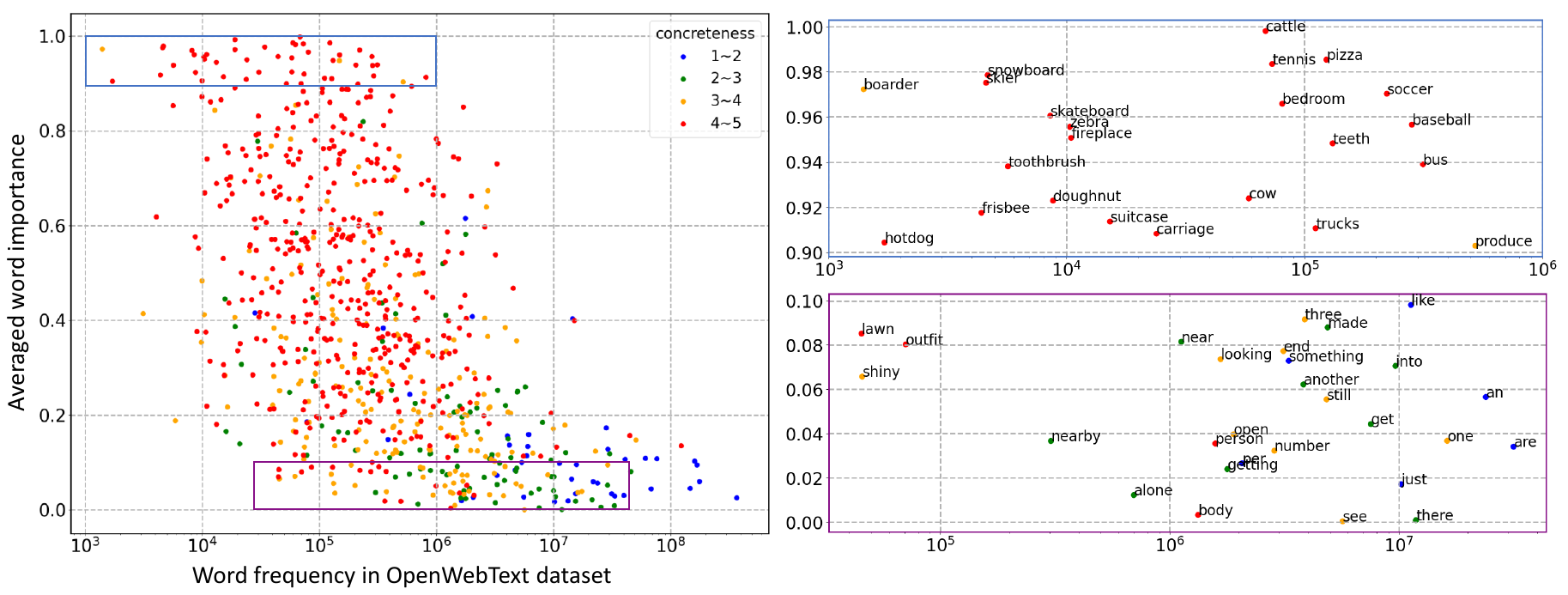}
	\end{center}
	\vspace{-0.3cm}
	\caption{The average word importance (via Grad-ECLIP) vs.~the word frequency in the OpenWordText dataset for the top-1000 most frequent words in the MS COCO Karpathy's validation split. The colors represent the concreteness level of each word according to \cite{brysbaert2014concreteness}. 
		(right) zoom-in of the blue and purple boxes to show the word examples with high word importance (blue box) and low word importance (purple box).
	}
	\label{fig:frequency}
	\vspace{-0.5cm}
\end{figure*}

\begin{figure}[th]
	\vspace{-0.1cm}
	\centering
	\begin{center}
		\includegraphics[width=0.4\textwidth]{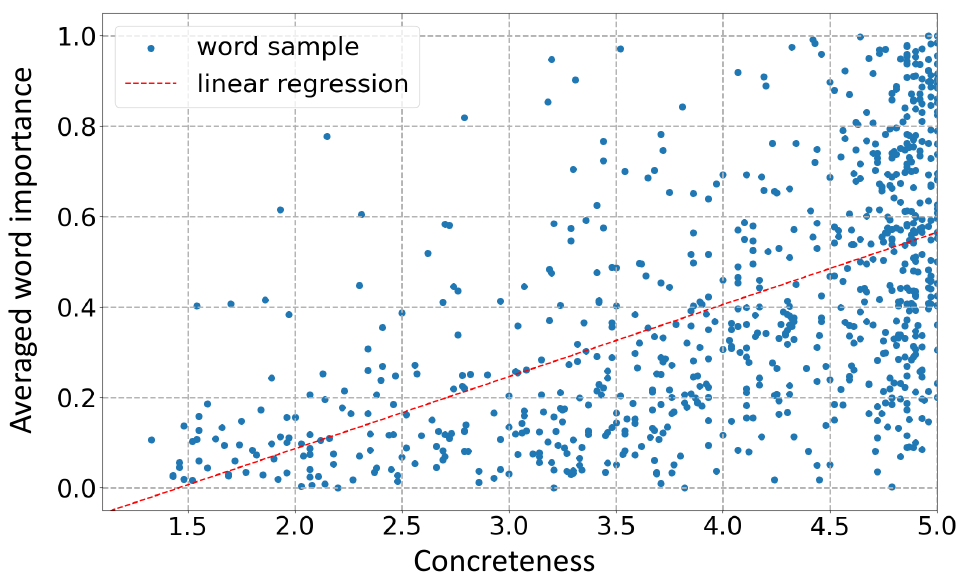}
	\end{center}
	\vspace{-0.3cm}
	\caption{The average word importance (via Grad-ECLIP) vs.~word concreteness for the top-1000 most frequent words in the MS COCO Karpathy's validation split. The red dashed line shows the linear regression result over the word samples; $r^2=0.32, p<0.001$.
	}
	\label{fig:concreteness}
	\vspace{-0.4cm}
\end{figure}

\vspace{-0.2cm}
\subsection{\zcy{Relationship between word importance and concreteness}}\label{sec:att_concreteness}
\vspace{-0.1cm}

From \S\ref{sec:diagn_attr}, 
we have inferred that the concrete \abc{visual} concepts \abc{(e.g., ``red'' and ``blue'')}  contribute more to the image and text matching than the abstract attributes \abc{(e.g., ``left'' and ``right'')}. Therefore, based on the text explanation obtained from Grad-ECLIP, we further explore the relationship between word importance in matching and word concreteness, and analyze which type of concepts and words (concrete vs. abstract) the CLIP model has actually learned and uses most often for matching. 
For an image-text pair, we calculate the textual explanation from Grad-ECLIP, and then obtain the importance value of each word by normalizing
such that the maximum word importance value is $1$ in the sentence. 
We then calculate the average word importance on the top-1000 most frequent words in the MS COCO caption (Karpathy's split) validation set.
For the word concreteness, we adopt the open-sourced database from \cite{brysbaert2014concreteness}, which provides the concreteness for 40,000 common English words measured by human rating. The concreteness is a value from 1 to 5, where 5 means the most concrete and 1 means the most abstract.

Based on the 1000 selected words, Fig. ~\ref{fig:concreteness} presents a scatter plot of average word importance versus concreteness value.
We perform linear regression analysis on this data (red dashed line), 
and the regression result was statistically significant ($r^2=0.32$, $p<0.001$).
\abc{The scatter distribution and linear regression result reveal that CLIP places higher word importance on more concrete words, and vice versa, less word importance on more abstract words. \textit{Therefore, the words that CLIP has learned for matching are biased towards concrete words.} 
}


\abc{CLIP's learned bias towards concrete words could be due to frequency bias in the training set, i.e., concrete words could appear more frequently during training.} 
To investigate this possibility, 
we attempt to count the number of occurrences of the selected words in CLIP's training set WebImageText \cite{radford2021clip}.
However, since WebImageText is not publicly available, 
we instead compute these statistics from the OpenWebText \cite{Gokaslan2019OpenWeb} dataset, which follows the same 
methodology 
to reproduce the data characteristics and structure of the WebImageText corpus. 

The relationship between average word importance and the word frequency in the training corpus is plotted in Fig.~\ref{fig:frequency}, with different sample colors 
representing the levels of concreteness.
There is no obvious relationship between the word attention and the frequency.
Many high-frequency words obtain a low word importance when the concreteness value is very low, and on the opposite, low-frequency words of concrete concepts may obtain a high word importance. 
Fig.~\ref{fig:frequency} (right) shows the zoom-in on the samples in the blue box, which have high average word importance, and in the purple box with low average importance. Comparing these two boxes, the words with concrete visual meaning, especially nouns, are indeed more likely to obtain higher importance than the abstract words. 
Therefore, the analysis using Grad-ECLIP text explanation further supports the conclusion from \S\ref{sec:diagn_attr} -- 
\textit{concrete visual concepts are learned better and contribute more to the CLIP image-text matching score than abstract concepts, and this phenomenon is not due to word frequency bias.} 



\vspace{-0.1cm}
\subsection{\rred{Limitations and future work}}
\vspace{-0.1cm}
\red{In this paper, we have analyzed CLIP via XAI methods by observing and interpreting the explanation heat maps on different samples,  such as in Sections V-I and V-II. However, it should be noted that there are limitations of heat map-based analysis, which stem from two aspects. Firstly, the observation results are illustrated through a small set of examples (compared to the size of the dataset), where limited samples will reduce the credibility of conclusions. 
Secondly, the faithfulness of any XAI method cannot reach a hundred percent, and the importance ranks obtained by certain methods cannot perfectly display both the necessity and sufficiency \cite{wang2020interpreting}. Therefore, conclusions drawn from XAI results from certain methods are subject to some extent.}
\rred{In future work, we will consider \red{when given image-sentence pair (with a sentence containing multiple words), how to associate each important word from the sentence to each important region} in the image, and vice versa. Future work can also consider how to extend Grad-ECLIP to other VLMs with modified dual-encoder architectures, e.g ALBEF with additional cross-attention layers to fuse image and text features.}

\vspace{-0.3cm}
\section{Conclusion}\label{sec:conclusion}
\vspace{-0.1cm}
In this paper, we propose Grad-ECLIP, a novel white-box gradient-based visual \abc{and textual} explanation method for CLIP, the dual-encoder pre-trained model for image-text matching. Grad-ECLIP is applicable to both the image and text encoders, producing heat maps that indicate the importance of image regions or words for the image-text matching score. Qualitative and quantitative evaluations demonstrate the advantages of Grad-ECLIP compared with existing explanation methods designed for transformers/CLIP, \zcy{and the adaptation experiments exhibit the generalizability of our method.} Finally, we also adopt Grad-ECLIP to analyze the properties of the pre-trained CLIP model, where we discover its ability of concept decomposition and addibility,  advantages/limitations on different attribute identification, \abc{and its bias towards learning concrete words over abstract words}. By introducing these analyses as examples, we hope the proposed interpretation method can be used to help with both development and understanding  of VLMs. 

\vspace{-0.3cm}
\section*{Acknowledgments}
This work was supported by Strategic Research Grants from City University of Hong Kong (Project. Nos. 7030010), and by Research Grant Council of Hong Kong (Project Nos AoE/E-601/24-N and T45-401/22N-4).
\vspace{-0.3cm}


%
\footnotesize
\bibliographystyle{IEEEtran}
\bibliography{references}


\vspace{-1cm}
\begin{IEEEbiography}[{\includegraphics[width=1in,height=1.25in,clip,keepaspectratio]{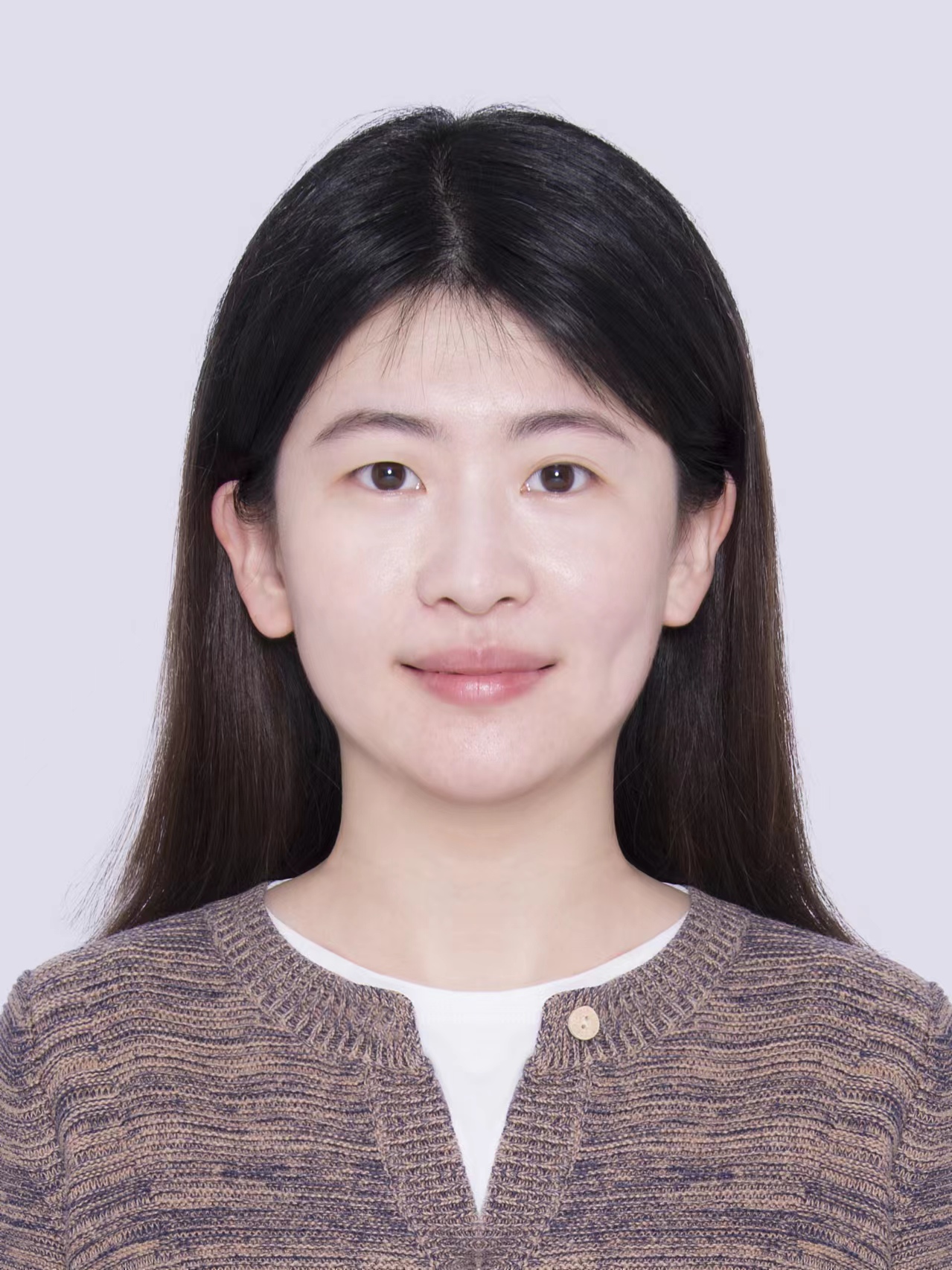}}]{Chenyang Zhao}
	received the B.Eng. degree in Electrical Engineering from Xiamen University, Xiamen, China, and M.S. degree in Computer Science from School of Electronic and Computer Engineering, Peking University, Shenzhen, China, in 2016 and 2019, respectively. She is currently working towards the Ph.D. degree in Computer Science at the City University of Hong Kong. Her research interests include explainable AI and object detection.
\end{IEEEbiography}

\vspace{-1cm}
\begin{IEEEbiography}[{\includegraphics[width=1in,height=1.25in,clip,keepaspectratio]{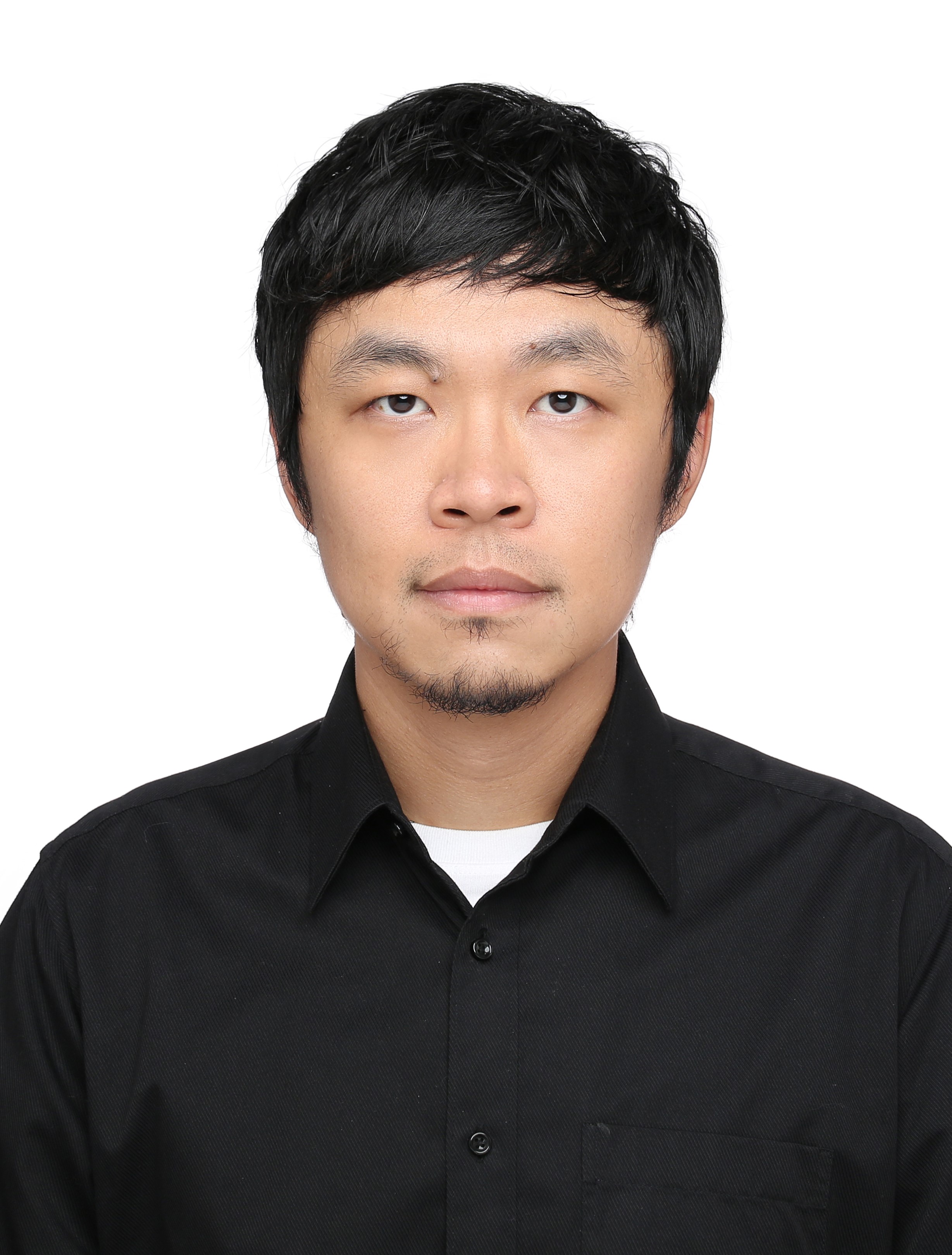}}]{Kun Wang}
	is a senior researcher at SenseTime Group Limited. He holds an MPhil degree from the Department of Electronic Engineering at the Chinese University of Hong Kong. His research interests focus on computer vision and representation learning. Presently, he is engaged in developing applications utilizing large language models and large multi-modal models within the industry.
	
\end{IEEEbiography}
\vspace{-1cm}
\begin{IEEEbiography}[{\includegraphics[width=1in,height=1.25in,clip,keepaspectratio]{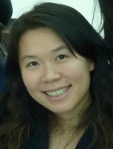}}]{Janet H. Hsiao}
	received the B.S. degree in Computer Science \& Information Engineering from National Taiwan University, the M.S. degree in Computing Science from Simon Fraser University, and the Ph.D. degree in Informatics from University of Edinburgh. She is currently a Professor in the Division of Social Science and Department of Computer Science \& Engineering at Hong Kong University of Science and Technology. She is also a Fellow of the Cognitive Science Society and serves on the Governing Board. Her research interests include cognitive science, computational modelling, learning and visual cognition, and explainable AI.
\end{IEEEbiography}

\vspace{-1cm}
\begin{IEEEbiography}[{\includegraphics[width=1in,height=1.25in,clip,keepaspectratio]{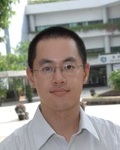}}]{Antoni B. Chan}
	received the B.S. and M.Eng. degrees in electrical engineering from Cornell University, Ithaca, NY, in 2000 and 2001, and
	the Ph.D. degree in electrical and computer engineering from the University of California, San Diego (UCSD), San Diego, in 2008. He is currently a Professor in the Department of Computer Science, City University of Hong Kong. His research interests include computer vision, machine learning, pattern recognition, and music analysis.
\end{IEEEbiography}

\vfill

\end{document}